\documentclass[11pt]{article}

\usepackage[margin=1in]{geometry}
\usepackage{setspace}
\usepackage{graphicx}
\usepackage{xparse}
\usepackage{amsmath, amssymb}
\usepackage{hyperref}
\usepackage{booktabs}
\usepackage{enumitem}
\usepackage{xcolor}
\usepackage{fancyhdr}
\usepackage{caption}

\NewDocumentCommand{\nbfigure}{O{0.9\textwidth} O{!tb} m m m}{%
  \begin{figure}[#2]
    \centering
    \includegraphics[width=#1]{figures/#3}
    \caption{#4}
    \label{fig:#5}
  \end{figure}%
}

\newcommand{\passhatk}[1][k]{%
  \text{pass\textasciicircum}\mkern1mu #1%
}

\hypersetup{
  colorlinks=true,
  linkcolor=blue,
  citecolor=blue,
  urlcolor=blue
}

\title{A Hierarchical Statistical Framework for Evaluating Agent Reliability}
\author{Mirco Milletarì and Tuhin Kundu \\Microsoft}
\date{\today}

\begin{document}
\maketitle

\begin{abstract}
  Agents operating in complex scenarios often face uncertainty due to incomplete information about the environment they operate in, ambiguous user queries, complex business rules and the inherently stochastic dynamics of the governing policy. Several new benchmarks~\cite{yao2024, huang-etal-2025-crmarena,huang-etal-2026-crmarena-pro,bandi2026mcpatlas, thinkingbox} have been proposed to evaluate LLM-based agents on complex tasks in multi-tool environments, including settings with policy constraints and varying levels of intent ambiguity. These conditions make it important to evaluate agentic systems systematically and quantify their performance under different conditions to assess their reliability in production. This need is evident in the recently introduced Thinkingbox benchmark~\cite{thinkingbox}, where even the strongest models struggle to perform consistently across diverse business scenarios. This paper presents a statistical framework for evaluating uncertainty and reliability in agentic systems. We define a hierarchy of scenarios, instances, and samples to assess model performance and variability systematically. We apply this framework to common metrics found in the literature: mean pass rate, $pass@k$~\cite{chen2021}, and $\passhatk$~\cite{yao2024}, with uncertainty estimates for each metric. Aggregating metrics at different levels of the hierarchy reveals different structures and properties of the policy model; we connect this hierarchical aggregation with the known process of \emph{coarse-graining} as intended in the physics and statistical literature. We independently derive exact finite-sample expressions for the sampling variances of the unbiased $pass@k$ and $\passhatk$ estimators; related results also appear in recent manuscripts~\cite{anonymous2026passk,anonymous2026rankings}. Our central contribution is a constructive derivation of hierarchical uncertainty propagation, making explicit its correspondence with coarse-graining and separating execution noise from task heterogeneity. Together, these derivations provide a unified statistical framework for evaluating reliability across repeated attempts, tasks, and scenarios.
\end{abstract}
%
\section{Introduction} \label{sec:introduction}
Large language models (LLMs) have enabled increasingly capable AI agents that combine a policy model, tools for interacting with an external environment, and a harness that orchestrates their execution. Given a task, an agent generates a solution trajectory comprising reasoning steps and actions, potentially modifying the environment along the way. Stochastic decision-making, task ambiguity, and complex interactions with the environment can produce different trajectories and outcomes across repeated executions of the same task. Evaluating these systems therefore requires more than a single aggregate performance score: we must establish how consistently they succeed, where failures concentrate, and how precisely performance has been measured.

Yet benchmark reporting often emphasizes a single state-of-the-art (SOTA) result without quantifying its statistical uncertainty~\cite{miller2024}. When reported differences between models amount to only a few percentage points, point estimates alone do not establish whether those differences reflect genuine improvements or sampling noise. This limitation is particularly consequential for agentic systems, whose performance can vary both across tasks and across repeated attempts at the same task. Bjarnason et al.~\cite{bjarnason2026} quantify this variability on SWE-Bench-Verified and show that small reported improvements can be comparable to evaluation noise. The recently introduced Thinkingbox benchmark~\cite{thinkingbox} likewise highlights substantial variability among strong models, both within and across domains, with uncertainty estimates complicating claims of a clear leader even within a single domain. These observations motivate a statistical treatment that distinguishes variability in agent behavior from uncertainty in benchmark estimates.

Reliability also extends beyond success frequency. Rabanser et al.~\cite{rabanser2026reliability} organize agent reliability into consistency, robustness, predictability, and safety, and find that capability gains need not yield comparable improvements in reliability. Their perspective is complementary to ours: rather than proposing a comprehensive taxonomy of reliability, we focus on the statistical interpretation of success-based metrics and the propagation of uncertainty across repeated attempts, tasks, and scenarios. Such analysis supports, but does not replace, evaluation of robustness or failure severity.

From a user's perspective, an agent should solve a task reliably without requiring repeated requests. Single-attempt success, summarized by $pass@1$ (mean pass rate), is therefore a central measure of practical performance. However, the average alone neither captures how consistently individual tasks are solved nor indicates how precisely performance has been estimated. Repeated evaluation of each task helps characterize execution variability, while confidence intervals quantify uncertainty in the resulting estimates. These measurements provide evidence relevant to production reliability, although their applicability depends on how closely the benchmark reflects deployment conditions.
Reliability must also be considered alongside the cost of obtaining a successful outcome. An agent may eventually find a solution through retries, but this behavior can be undesirable for long-horizon tasks that consume substantial time and computational resources. Secondary metrics such as number of tool calls, retries, and token usage therefore complement success-based measures by characterizing the efficiency of solution trajectories. Together, these measurements can guide reinforcement fine-tuning (RFT) toward more consistent and efficient task completion, rather than improvements in aggregate success alone.

Prior work provides complementary approaches to statistical evaluation. Chen et al.~\cite{chen2021} introduce the unbiased $pass@k$ estimator, while Yao et al.~\cite{yao2024} introduce $\passhatk$ and its unbiased estimator to measure agent reliability across repeated trials. Hariri et al.~\cite{hariri2026} develop a Bayesian alternative to $pass@k$-based model ranking, with analytical posterior uncertainty and credible intervals for underlying success probabilities. Dewasurendra~\cite{dewasurendra2026} studies a different problem: finite-sample confidence intervals and identification limits for population $pass@k$, particularly when extrapolating beyond the observed sampling budget. These questions are distinct from the exact frequentist sampling variances of the unbiased $pass@k$ and $\passhatk$ estimators studied here; bootstrap uncertainty estimates likewise do not provide analytical expressions for those variances. Recent manuscripts also analyze finite-sample uncertainty for these metrics~\cite{anonymous2026passk,anonymous2026rankings}, but do not address hierarchical uncertainty propagation.

In this work, we present a unified statistical framework for evaluating performance and uncertainty in LLM-based agentic systems. Our contributions are threefold. First, we organize evaluation into scenarios, task instances, and repeated samples, distinguishing execution variability, differences across tasks, and uncertainty in estimated performance. We interpret aggregation through this hierarchy as \emph{coarse-graining} in physics and statistics~\cite{kardar2007}, showing how averages reveal collective behavior while obscuring task-level differences. Second, we analyze uncertainty in mean pass rate and derive exact finite-sample sampling variances for the unbiased $pass@k$ and $\passhatk$ estimators, connecting these per-task results to uncertainty in population-level averages. Third, we illustrate the framework on agentic benchmarks to show what these metrics and their uncertainty estimates reveal about reliability. The central contribution is not an isolated metric or variance formula, but an account of how task heterogeneity and finite evaluation budgets jointly shape the conclusions drawn from benchmark results. The framework draws on established statistical results; the coarse-graining connection and analytical derivations are developed in Appendices~\ref{ch:metrics-coarse-graining} and \ref{ch:pass-k-cumulants}.
\section{Modeling Assumptions and Definitions} \label{sec:Modeling-Approach}
Throughout the paper, the policy model governs the agent's decisions, while the environment includes the tools and infrastructure with which it interacts. We evaluate a fixed agent configuration, including its policy, harness, and environment setup. Following the terminology used in \cite{thinkingbox}, we define the hierarchical structure of our framework as follows:
\begin{enumerate}[leftmargin=*]
  \item \textbf{Scenario}: A scenario is defined by a specific set of tools (\textit{e.g.}, a set of MCP servers) and the environment's initial state. It is grounded in the real-world context it aims to simulate, such as retail, healthcare, finance, etc.
  \item \textbf{Instance}: A specific task that the agent must complete using the tools available in a given scenario. An instance is a data point in our dataset comprising the user's intent, the system prompt, and other relevant context. A dataset is a collection of instances across one or more scenarios.
  \item \textbf{Sample}: A single attempt by the agent to resolve a given instance, also referred to as a "sample trajectory." To assess the agent's performance and variability, we need multiple samples for each instance.
\end{enumerate}
\nbfigure[\textwidth][!tb]{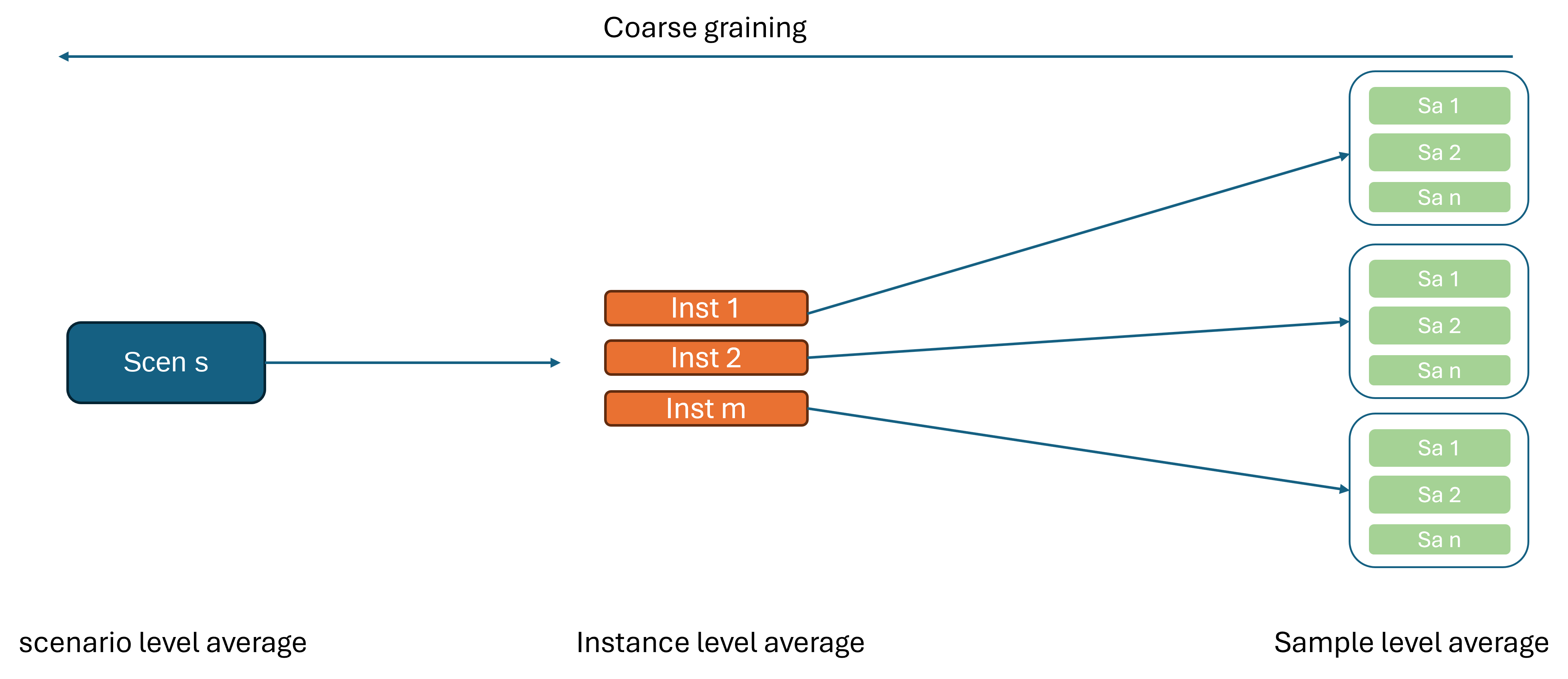}{
  Hierarchical data structure: scenarios contain multiple instances, each instance is sampled multiple times to account for variability. Aggregating metrics at different levels of this hierarchy corresponds to coarse graining.
}{data-hierarchy}
For the analysis, we write $\pi_\theta(y\mid x)$ for the trajectory distribution induced by the policy with parameters $\theta$ in this fixed evaluation setup. Here, $x$ specifies the task and its context, and $y$ is the resulting trajectory; the distribution includes any modeled environment or user-interaction randomness.
Conceptually, population evaluation involves two levels of sampling: a task $x$ is drawn from an unknown population distribution $P_{\mathrm{task}}$, and a trajectory $y$ is then generated conditionally on that task:
\begin{equation}  
  x \sim P_{\mathrm{task}},
  \qquad
  y \mid x \sim \pi_\theta(\cdot \mid x).
\end{equation}
In practice, we evaluate a finite dataset $\mathcal{D}=\{x_1,\ldots,x_{|\mathcal{D}|}\}$ and repeatedly execute each task. Section~\ref{ssec:task-success-distributions} connects this dataset to the task population and the agent's task-specific success probabilities.

In the agentic context, the final output $y$ comprises a sequence of actions generated by the policy model and executed by the orchestrator in the environment, and may also include multi-turn interactions with the user (see Fig.~\ref{fig:trajectory}). We define a sample trajectory as follows:
\nbfigure[\textwidth][!tb]{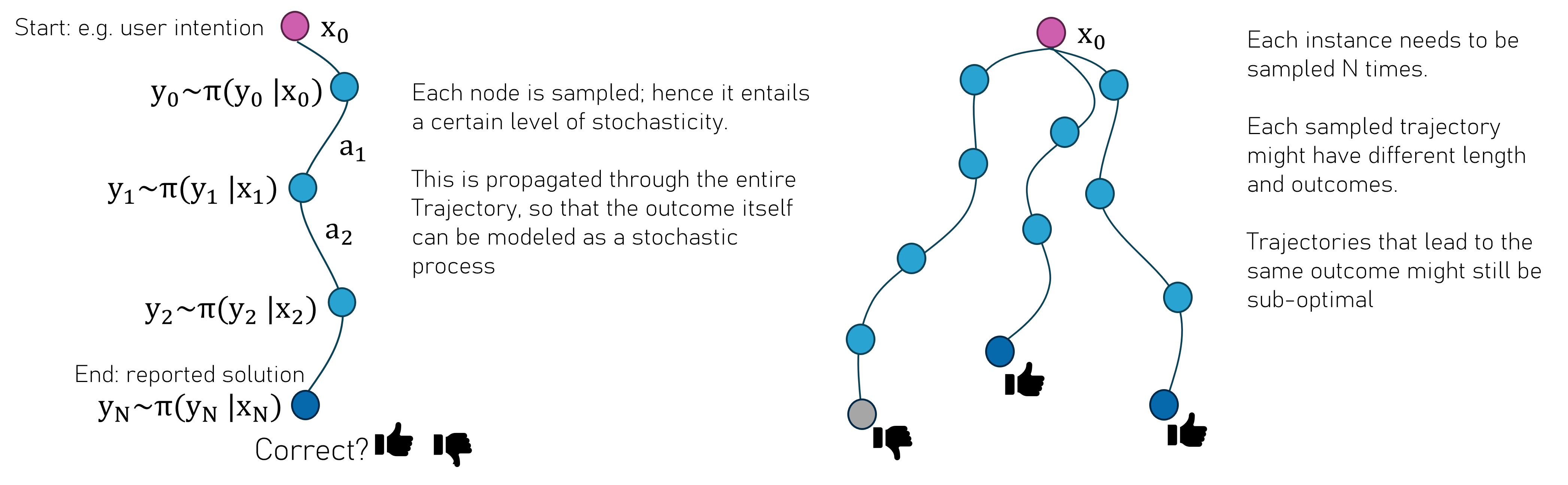}{
  Sample trajectory: (left) a single attempt of the agent to resolve a given instance, comprising actions, states, and user interactions.
  (right) Repeatedly sampling the same instance typically leads to different solution paths and possibly different outcomes.
}{trajectory}
\begin{equation} \label{eq:trajectorydef}
  \hat{Y} = \{A, S, U\},
\end{equation}
where $A$ is the sequence of actions taken by the model, $S$ is the sequence of states observed in the environment, and $U$ is the sequence of user interactions (if any)~\footnote{Note that with $A$ we denote the sequence-level action, e.g. calling a tool, reasoning, rather than the next-token generation from a prefix, tipically denoted as $a$.}. An evaluation experiment can be understood as a rollout of the policy model $\pi_\theta$ on a set of instances, each associated with its own evaluation criterion. As in any experiment, stochasticity arises from various factors. We account for it by running multiple samples per instance and statistically analyzing the results to assess the model's average performance and variability.

Throughout the paper, we adopt the following modeling assumption: each sample outcome is a random variable $\xi_{s,i,r}$, where $s$ is the scenario index, $i$ is the instance index, and $r$ is the sample index. For this analysis, we consider binary outcomes: graders report each experiment as a success or a failure according to whether the task meets a specified set of evaluation criteria. The hierarchical variance decomposition extends to other rewards with finite variance, but the Bernoulli and success-metric formulas below rely on binary outcomes. Under this model, each sample is drawn from a Bernoulli distribution conditional on the true success probability $\rho_i$ of instance $i$:
\begin{equation} \label{eq:stochdef}
  \xi_{s,i,r} | \rho_i \sim \text{Bernoulli}(\rho_i)
\end{equation}
Here, $\rho_i$ is the true success probability for instance $i$. In practice, we estimate it by running the model $N$ times on that instance and computing the empirical pass rate $p_i=C_i/N$, where $C_i$ is the number of successful samples. Instances differ in their probability of success, and variation among the resulting estimates reflects both differences between instances and finite-sample noise. The statistical framework below separates these contributions.

Intuitively, evaluation proceeds in two steps: repeated attempts estimate each instance's pass rate, then averaging these estimates gives a scenario-level mean pass rate. Section~\ref{ssec:pass-rate-undertainty} explains how uncertainty propagates through these averages when estimating performance on a broader task population.
\begin{table}[!t]
  \centering
  \begin{tabular}{@{}ll@{}}
    \toprule
    \textbf{Metric} & \textbf{Main intuition} \\
    \midrule
    \textbf{Pass@1} & Typical single-attempt performance \\
    \textbf{Pass@k} & Can retries rescue the task? \\
    \textbf{Pass\textasciicircum k} & Is success consistently reproducible? \\
    \bottomrule
  \end{tabular}
\end{table}
From a practical perspective, a task's estimated success probability indicates how likely an attempt is to succeed, while its uncertainty estimate indicates how precisely that probability is known. Other metrics considered here include \emph{pass@k} and \emph{$\passhatk$}. For a given task, the former measures the probability that at least one of $k$ independent attempts succeeds; the latter measures the probability that all $k$ attempts succeed.
\section{Statistical Analysis Framework}\label{sec:statistical-analysis-framework}
The dataset's hierarchical structure (scenarios, instances, and samples) requires global metrics that capture model performance at different levels of granularity. Moving up the hierarchy corresponds to a \textit{coarse-graining} process in which we aggregate performance metrics by averaging. However, this process can obscure important information about model performance in specific domains and on individual instances. For example, Ref.~\cite{thinkingbox} presents a detailed benchmark of several models at the scenario and dataset levels and shows that no model consistently outperforms the others across all scenarios, even when some models lead in the across-scenario rankings. This finding highlights the importance of considering both fine- and coarse-grained metrics. Although pass rates can be aggregated throughout the hierarchy by simple averaging, propagating uncertainty across levels requires additional care, particularly for evaluations with limited numbers of instances~\cite{miller2024,bowyer2025}. We address this issue in Section~\ref{ssec:pass-rate-undertainty} within the hierarchical uncertainty-propagation framework, and provide an explicit derivation in Appendix~\ref{ch:metrics-coarse-graining}, which accounts for variability at multiple levels and their interactions; the derivation borrows the formalism of~\emph{coarse-graining}~\cite{kardar2007} from the physics and statistical literature. Informally, coarse-graining is the process of zooming out the description of a system, moving from a microscopic to a macroscopic description, e.g. describing a fluid in terms of its individual constituents versus a continuous field. Mathematically, zooming out consists in averaging the microscopic variables, which inevitably leads to a loss of resolution while highlighting collective, population-level aspects.
\subsection{From task distributions to success probabilities}\label{ssec:task-success-distributions}
A benchmark contains tasks, not success probabilities. The connection is that each task has a probability of success for the particular agent being evaluated. Let $r(x,y)\in\{0,1\}$ indicate whether trajectory $y$ successfully completes task $x$ according to its evaluation criterion. For a fixed agent configuration and grader, define
\begin{equation} \label{eq:task-prob}
  \rho_\theta(x) = \mathbb{E}_{y\sim\pi_\theta(\cdot\mid x)}[r(x,y)],
  \qquad
  \rho_i \equiv \rho_\theta(x_i).
\end{equation}
For a fixed task, $\rho_\theta(x)$ is a fixed, generally unknown number. Repeated attempts randomize the outcome, not this underlying probability. A benchmark author may anticipate difficulty from task characteristics such as the required reasoning or tool dependencies. However, success probability measures difficulty relative to the evaluated agent and grader: another agent may succeed more or less often on the same task. Like a school assignment, a task can have an intended level of complexity while posing different challenges to different students.

Now imagine selecting tasks from $P_{\mathrm{task}}$ and recording each selected task's true success probability. The resulting distribution of probabilities is $g$. It describes differences across tasks, not uncertainty about the probability of one fixed task. For example, suppose a population contains two tasks, selected with probabilities $0.75$ and $0.25$, on which the agent succeeds with probabilities $0.2$ and $0.8$, respectively. Then $g$ assigns weight $0.75$ to $0.2$ and weight $0.25$ to $0.8$. The weights come from how often tasks are selected; the success probabilities come from the agent's performance on those tasks. Changing either the task mix or the agent can therefore change $g$. Formally, this transformation of the task distribution is called its \emph{pushforward} through $\rho_\theta$~\cite{tao2011,villani2003,santambrogio2015}:
\begin{equation} \label{eq:pushforward}
  g=(\rho_\theta)_\#P_{\mathrm{task}},
  \qquad
  F_g(t)=\Pr_{x\sim P_{\mathrm{task}}}\!\left(\rho_\theta(x)\le t\right).
\end{equation}
Here, $F_g(t)$ is the fraction of the task population, weighted by its selection probabilities, on which the agent's success probability is at most $t$. Writing $\rho\sim g$ is shorthand for selecting a task and taking its associated success probability, not an additional independent sampling step:
\begin{equation} \label{eq:sampling-hierarchy}
  x\sim P_{\mathrm{task}}
  \quad\longrightarrow\quad
  \rho=\rho_\theta(x)
  \quad\longrightarrow\quad
  \xi\mid\rho\sim\operatorname{Bernoulli}(\rho).
\end{equation}
The population's single-attempt success probability is therefore the mean of $g$: $\Pr(\xi=1)=\mathbb{E}_{x\sim P_{\mathrm{task}}}[\rho_\theta(x)]=\mathbb{E}_{\rho\sim g}[\rho]$. In the two-task example, it is $0.75\times0.2+0.25\times0.8=0.35$. For the finite dataset $\mathcal D$, selecting an equally weighted task means using the empirical task distribution
\begin{equation}
  \widehat P_{\mathcal D}
  =\frac{1}{|\mathcal D|}\sum_{i=1}^{|\mathcal D|}\delta_{x_i},
\end{equation}
where $\delta_{x_i}$ is a unit point mass at task $x_i$. This describes selection from the benchmark; it does not make the population distribution known. Replacing each task by its true success probability gives
\begin{equation} \label{eq:finite-dataset-distribution}
  g_{\mathcal D}
  =(\rho_\theta)_\#\widehat P_{\mathcal D}
  =\frac{1}{|\mathcal D|}\sum_{i=1}^{|\mathcal D|}
    \delta_{\rho_\theta(x_i)}.
\end{equation}
Thus, $\widehat P_{\mathcal D}$ puts equal weight on the observed tasks, while $g_{\mathcal D}$ puts the same weight on their true success probabilities. The mean of $g_{\mathcal D}$ is the true dataset-level pass rate, $|\mathcal D|^{-1}\sum_i\rho_i$. In practice, the $\rho_i$ are unknown. Replacing them by the observed pass rates $p_i=C_i/N$ defined in Section~\ref{sec:Modeling-Approach} gives
\begin{equation} \label{eq:empirical-distribution}
  \widehat g_{\mathcal D}
  =\frac{1}{|\mathcal D|}\sum_{i=1}^{|\mathcal D|}\delta_{p_i}.
\end{equation}
Unlike $g_{\mathcal D}$, $\widehat g_{\mathcal D}$ contains finite-run estimation noise as well as differences in true success probabilities. Even tasks with identical true probabilities can have different observed pass rates after a limited number of attempts. For a single binary outcome, the true success probability determines its distribution, but an estimated pass rate alone does not tell us how precisely that probability is known. Likewise, the dataset's mean pass rate alone does not reveal how success probabilities vary across tasks.

Finally, estimating performance on the fixed benchmark and estimating performance on a broader task population are different goals. For the benchmark's own mean, more independent attempts can make estimation uncertainty arbitrarily small. For the population mean, even perfectly known probabilities on those same tasks leave uncertainty about unobserved tasks. The analysis below models benchmark tasks as independent draws from a common task population to quantify this second contribution. This is a modeling assumption, not a guarantee of benchmark construction: deliberately selected or related tasks may require a different sampling model, and more attempts cannot correct an unrepresentative task mix.
\subsection{Pass rate and hierarchical uncertainty propagation} \label{ssec:pass-rate-undertainty}
In this section we estimate the scenario's population mean success probability using the law of total variance. We first compute per-instance pass rates and associated uncertainties, then aggregate them at the scenario level by averaging across independently sampled instances.

Uncertainty propagation is not trivial, however, because it encompasses both within-instance variability (due to stochasticity in model behavior) and between-instance variability (due to differences in instance difficulty).
Consider the first level of coarse-graining, from instances to a scenario:
\begin{enumerate}[leftmargin=*]
  \item Within a fixed instance, the $N$ outcomes are i.i.d. Bernoulli conditional on its true success probability $\rho_i$. The number of successes is $C_i$, and the empirical pass rate is $p_i$:
    \begin{equation}
      C_i \mid \rho_i \sim \operatorname{Binomial}(N, \rho_i),
      \qquad
      p_i = \frac{C_i}{N},
      \qquad
      \sigma^2_{p_i} := \operatorname{Var}(p_i \mid \rho_i) = \frac{\rho_i(1-\rho_i)}{N}.
    \end{equation}

  \item Across instances, the success probabilities vary with task difficulty. We model them as independent and identically distributed draws from a common, unknown distribution $g$:
    \begin{equation}
      \rho_i \sim g, \qquad \mathbb{E}_g[\rho] = \mu, \qquad \operatorname{Var}_g(\rho) = \tau^2.
    \end{equation}
    Here, $g$ is induced by the task distribution within the scenario under consideration through the map $\rho_\theta$ defined above. We only require that the first two moments of $g$ exist and are finite.

  \item Conditional on their success probabilities, the runs are independent within and across instances. The instance probabilities $\rho_i$ are also independent across instances.
\end{enumerate}
Now consider $M=|S|$ instances in scenario $S$. The scenario-level empirical pass rate is the average of the instance-level pass rates:
\begin{equation} \label{eq:scenario-pass-rate}
  \bar{p} = \frac{1}{M} \sum_{i=1}^M p_i.
\end{equation}
Its uncertainty follows from the law of total variance (see, e.g., Ref.~\cite{variance-wiki} and the derivation in Appendix~\ref{ch:metrics-coarse-graining}):
\begin{equation} \label{eq:total-variance}
  \operatorname{Var}(\bar p) = \mathbb{E}\!\left[\operatorname{Var}(\bar p \mid \rho_{1:M})\right]
  + \operatorname{Var}\!\left(\mathbb{E}[\bar p \mid \rho_{1:M}]\right).
\end{equation}
Conditional on $\rho_{1:M}$, the instance-level pass rates are independent, so
\begin{equation}
  \operatorname{Var}(\bar p \mid \rho_{1:M})
  = \frac{1}{M^2} \sum_{i=1}^M \operatorname{Var}(p_i \mid \rho_i)
  = \frac{1}{M^2} \sum_{i=1}^M \sigma^2_{p_i}.
\end{equation}
Averaging over the instance distribution gives the expected within-instance contribution:
\begin{equation}
  \mathbb{E}\!\left[\operatorname{Var}(\bar p \mid \rho_{1:M})\right]
  = \frac{1}{M^2} \sum_{i=1}^M \mathbb{E}_g[\sigma^2_{p_i}]
  = \frac{\mathbb{E}_g[\sigma^2_{p_i}]}{M}.
\end{equation}
The conditional mean is
\begin{equation}
  \mathbb{E}[\bar p \mid \rho_{1:M}] = \frac{1}{M} \sum_{i=1}^M \rho_i.
\end{equation}
Since the $\rho_i$ are i.i.d. with variance $\tau^2$, the between-instance contribution is
\begin{equation}
  \operatorname{Var}\!\left(\mathbb{E}[\bar p \mid \rho_{1:M}]\right)
  = \operatorname{Var}\!\left(\frac{1}{M} \sum_{i=1}^M \rho_i\right)
  = \frac{\tau^2}{M}.
\end{equation}
Combining the two terms gives the total variance of the scenario-level pass rate~\cite{miller2024}:
\begin{equation} \label{eq:total-variance-final}
  \operatorname{Var}(\bar p)
  = \frac{\tau^2 + \mathbb{E}_g[\sigma^2_{p_i}]}{M}
\end{equation}
The first term is between-instance variability and is unaffected by increasing the number of runs $N$; this term would still exist when $N=1$. The second term is within-instance sampling variability and decreases as $N$ increases. Both terms decrease as the number of independently sampled instances $M$ increases. Note that if the different tasks had the same difficulty, meaning the same pass-rate probability, the extra term $\tau^2=0$ and only sampling noise would be relevant.

Although run-level variability is averaged as we move up the hierarchy, it remains important at the instance level, where it can identify areas for model improvement and is closely tied to user experience. The same decomposition applies at the next level of the hierarchy, from scenarios to the global level, by treating scenarios as the units being averaged. Applying it recursively yields global performance metrics and uncertainty estimates that account for variability at multiple levels. 

To summarize, each instance is sampled $N$ times to capture within-instance sampling variability; for each instance we can estimate its pass rate and the standard-error on it. The second step, we take an average across instances, that is the average of the pass rates. This \emph{mean pass rate} now has an error propagated over from each individual instance, and one that captures the variability across instances, that is a population level variability, not a single-instance variability. The two errors scale differently with $N$ and $M$ and reveal different aspects of the agent's performance on the eval set. In the next section we use a toy model as a proxy to show this difference explicitly. 
\subsubsection{A "random matrix" toy model}
To make the two averaging steps explicit, consider a toy model with an $N\times M$ data matrix $\hat X$. Columns represent tasks and rows represent repeated attempts. We retain $i=1,\ldots,M$ as the instance index and $r=1,\ldots,N$ as the run index, writing the matrix entry as $X_{ri}$; for $l=1$, this corresponds to $\xi_{ir}$ above.

In a real benchmark, each column is a task $x_i$ whose agent-specific success probability $\rho_\theta(x_i)$ is unknown. Here, drawing a probability from $g$ stands in for sampling a task and taking its associated probability, preserving the binary-outcome model under the assumptions of Section~\ref{ssec:task-success-distributions} without generating task content. The model is
\begin{equation}
    \rho_1,\ldots,\rho_M \overset{\mathrm{iid}}{\sim} g, \quad X_{ri}\mid \rho_1,\ldots, \rho_M \sim \operatorname{Binomial}(l,\rho_i),
\end{equation}
where entries are conditionally independent given the task probabilities. Draw the $M$ probabilities once, then generate $N$ entries in each column using that column's probability. The positive integer $l$ counts Bernoulli trials per entry, with $l=1$ recovering binary outcomes. The overall average is
\begin{equation} \label{eq:Xav}
    \bar X=\frac{1}{N M}\sum_{i=1}^{M}\sum_{r=1}^{N}X_{ri}.
\end{equation}
For $l=1$, this is equivalent to Eq.~\eqref{eq:scenario-pass-rate}, with the two-level average making the hierarchy explicit: repeated entries within each column estimate one task's mean performance, while differences across column means reflect both task heterogeneity and finite-run noise. Averaging the column means gives the same point estimate as averaging all entries, but uncertainty propagation must preserve the task structure. As in the preceding subsection, the target is population performance: the unconditional variance considers variation in both the selected tasks and their execution outcomes. The conditional moments are
\begin{equation} \label{eq:unconditional-moments}
    \mathbb E[X_{ri}\mid \rho_i]=l \, \rho_i, \quad \mathbb E[X_{ri}^2\mid \rho_i] =l \, \rho_i(1-\rho_i)+l^2 \, \rho_i^2.
\end{equation}
From here, the unconditional moments can be defined using the shorthand notations:
\begin{equation} \label{eq:conditional-qv}
        q = \mathbb{E}_{\rho \sim g} [\rho (1-\rho) ] , \quad v = \operatorname{Var}_{\rho \sim g}(\rho). 
\end{equation}
In the notation of the preceding subsection, $v=\tau^2$ and $q/N=\mathbb{E}_g[\sigma^2_{p_i}]$, where $\sigma^2_{p_i}$ is the conditional sampling variance of the Bernoulli pass-rate estimator. Thus, for $l=1$, $v$ and $q/N$ we recover the between-instance and expected within-instance variance contributions, respectively.
Using this in the definition of the unconditional variance of $X_{ri}$:
\begin{equation} \label{eq:varX}
    \operatorname{Var}(X_{ri}) = \mathbb{E}[X^2_{ri}] - \mathbb{E}[X_{ri}]^2 = l \, q + l^2 \, \mathbb{E}[\rho^2]-l^2 \mathbb{E}[\rho]^2 
    = l \, q + l^2 \, v.
\end{equation}
\begin{figure}[!tb]
  \centering
  \begingroup
  \small
  \textbf{(a) Data: rows are repeated samples; columns are instances}
  \[
    \begin{array}{c@{\quad}c}
      & \begin{array}{cccc}
          \rho_1 & \rho_2 & \cdots & \rho_M
        \end{array}\\[-2pt]
      \hat X= &
      \begin{pmatrix}
        X_{11} & X_{12} & \cdots & X_{1M}\\
        X_{21} & X_{22} & \cdots & X_{2M}\\
        \vdots & \vdots & \ddots & \vdots\\
        X_{N1} & X_{N2} & \cdots & X_{NM}
      \end{pmatrix}
    \end{array}
  \]
  The overall average is defined in Eq.~\eqref{eq:Xav}.

  \textbf{(b) Covariance within one column: pair every row with every row}
  \[
    \Sigma_i=
    \begin{pmatrix}
      \textcolor{blue}{d} & \textcolor{red!70!black}{o} & \cdots & \textcolor{red!70!black}{o}\\
      \textcolor{red!70!black}{o} & \textcolor{blue}{d} & \cdots & \textcolor{red!70!black}{o}\\
      \vdots & \vdots & \ddots & \vdots\\
      \textcolor{red!70!black}{o} & \textcolor{red!70!black}{o} & \cdots & \textcolor{blue}{d}
    \end{pmatrix},
    \qquad
    \begin{array}{ll}
      \textcolor{blue}{d=lq+l^2v}: & N\text{ diagonal entries},\\[3pt]
      \textcolor{red!70!black}{o=l^2v}: & N(N-1)\text{ off-diagonal entries}.
    \end{array}
  \]
  \endgroup
  \caption{Data layout and unconditional covariance structure, panels (a)--(b). Each data column shares one latent probability $\rho_i$. Diagonal covariance entries contain both sampling noise and instance variability; off-diagonal entries within a block contain only the shared instance contribution.}
  \label{fig:matrix-covariance-structure}
\end{figure}
With this, we can compute the unconditional variance of $\bar{X}$ using Eq.\eqref{eq:Xav}:
\begin{align} \label{eq:varXav}
    \operatorname{Var}(\bar{X}) = \frac{1}{N^2 M^2} \operatorname{Var}\left(\sum_{i=1}^{M}\sum_{r=1}^{N} X_{ri} \right) = \frac{1}{N^2 M^2} 
    \sum_{i=1}^{M} \sum_{r=1}^{N}\sum_{r'=1}^{N} \operatorname{Cov}(X_{ri},X_{r'i}).
\end{align}
Figure~\ref{fig:matrix-covariance-structure} distinguishes the data matrix from the covariance matrix used in this sum. For a fixed column $i$, let $\mathbf{x}_i=(X_{1i},\ldots,X_{Ni})^\top$; its unconditional covariance matrix has entries
\begin{equation}
    (\Sigma_i)_{rr'}
    = \operatorname{Cov}(X_{ri},X_{r'i})
    = lq\,\delta_{rr'}+l^2v.
\end{equation}
For $r\ne r'$, conditional independence gives $\mathbb{E}[X_{ri}X_{r'i}\mid\rho_i]=l^2\rho_i^2$, hence the off-diagonal covariance is $l^2v$. This unconditional covariance arises because both entries share the same randomly selected task probability; the executions remain independent conditional on that probability. Different columns are independent and have zero cross-covariance. The diagonal here therefore refers to pairing an entry with itself, not to the diagonal of the rectangular data matrix $\hat X$.
\begin{figure}[!tb]
  \centering
  \begingroup
  \small
  \textbf{All columns: independent covariance blocks}
  \[
    \operatorname{Cov}
    \begin{pmatrix}\mathbf{x}_1\\ \mathbf{x}_2\\ \vdots\\ \mathbf{x}_M\end{pmatrix}
    =
    \begin{pmatrix}
      \Sigma_1 & 0 & \cdots & 0\\
      0 & \Sigma_2 & \cdots & 0\\
      \vdots & \vdots & \ddots & \vdots\\
      0 & 0 & \cdots & \Sigma_M
    \end{pmatrix}
  \]
  \[
    \operatorname{Var}(\bar X)
    =
    \frac{
      \overbrace{MN}^{\text{diagonal count}}\,
      \textcolor{blue}{d}
      +
      \overbrace{MN(N-1)}^{\text{off-diagonal count}}\,
      \textcolor{red!70!black}{o}
    }{(NM)^2}
    =\frac{lq}{NM}+\frac{l^2v}{M}.
  \]
  \endgroup
  \caption{Unconditional covariance across all task columns. Blocks for different columns have zero cross-covariance. The variance of the overall average is the sum of all covariance entries divided by $(NM)^2$.}
  \label{fig:matrix-covariance-blocks}
\end{figure}
Fig.~\eqref{fig:matrix-covariance-blocks} shows the covariance structure across all task columns. Each of the $M$ covariance blocks has $N$ diagonal elements with $l q+l^2 v$ and $N(N-1)$ ordered off-diagonal elements with $l^2 v$. Using this in Eq.~\eqref{eq:varXav} we obtain the general expression for the unconditional variance:
\begin{equation} \label{eq:xvar-final}
    \operatorname{Var}(\bar X) = \frac{M}{N^2 M^2} \left[N (l q+l^2 v)+ N(N-1) l^2 v\right]
     = \frac{lq}{N M}+\frac{l^2 v}{M}.
\end{equation}
Choosing $l=1$, we now recover Eq.~\eqref{eq:total-variance-final}. More rows improve the precision of each task mean, reducing the within-task contribution by $1/N$. The between-task contribution reflects variation in true success probabilities and does not receive this additional reduction. Both contributions are reduced by averaging over $M$ tasks. If every task has the same true success probability, $v=0$; sampling different probabilities from a common $g$ does not impose that condition.
\nbfigure[0.9\textwidth][!tb]{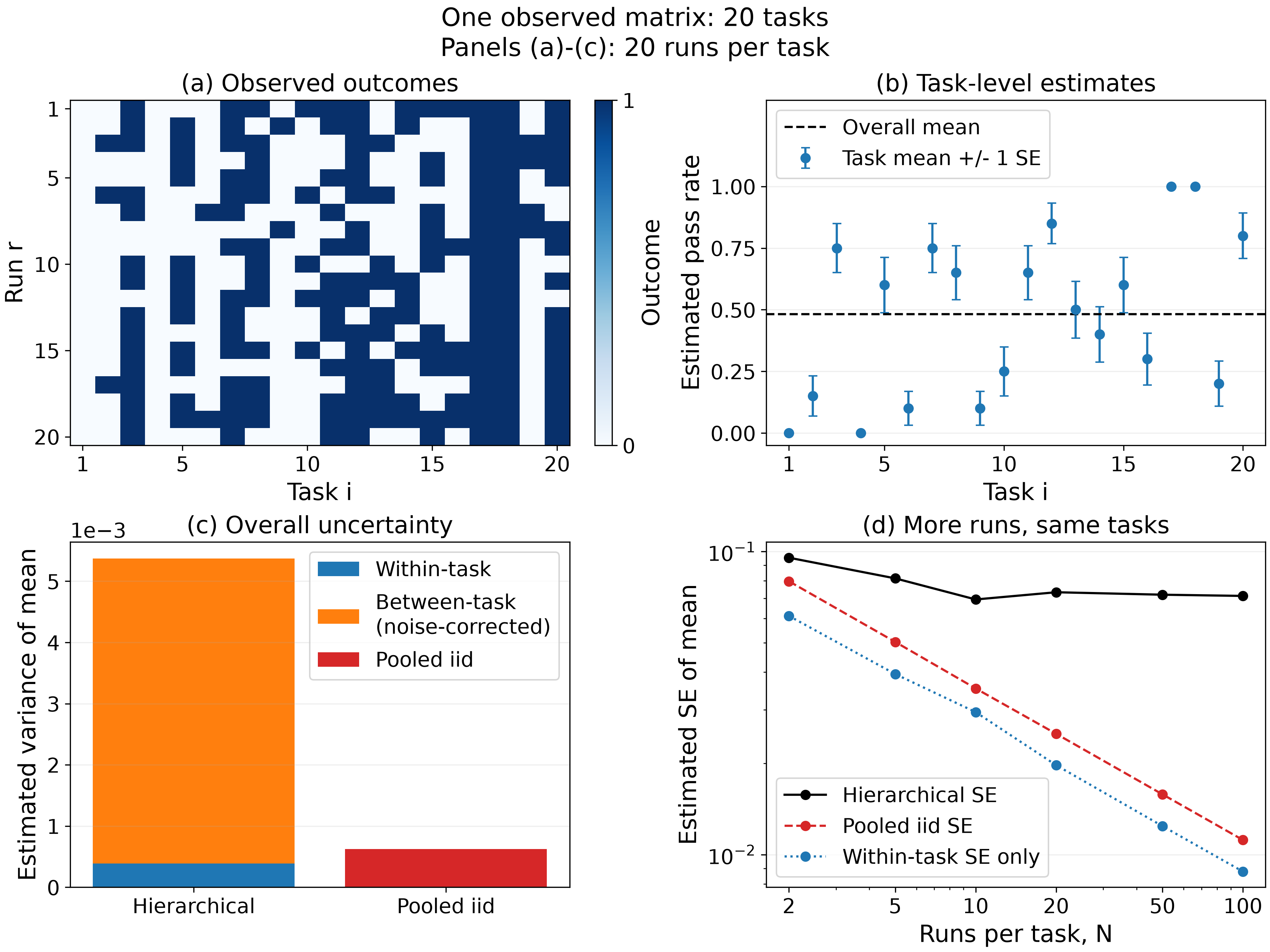}{
  Two-step averaging of one observed matrix with $M=20$ tasks. (a) The outcome of the matrix generation process for the $M$ tasks, showing only the first $N=20$ rows of binary outcomes for convenience. (b) Task-level means with estimated SE. (c) The within-task and noise-corrected between-task contributions to the hierarchical variance estimate, compared with the pooled estimate; both methods give the same mean. (d) SE estimates from nested row prefixes of the same $100\times20$ matrix. Repeated attempts are not additional independent tasks.
}{hierarchical-aggregation}
\paragraph{An explicit numerical example}

To make things concrete, we compare two ways of estimating uncertainty in population mean performance from the same observations: preserving the task structure, or incorrectly treating every execution as an independent task. Both produce the same mean, but different standard errors.
Figure~\ref{fig:hierarchical-aggregation} follows the analysis of a single observed matrix with $M=20$ tasks and binary outcomes ($l=1$). The task probabilities $\rho_i$ are drawn once from $g=\operatorname{Uniform}(0,1)$, then for each column $i$, we sample $N=100$ entries independently with success probability $\rho_i$ to obtain a $100 \times 20$ matrix. Eq.~\eqref{eq:Xav} can be though of as a two step process: first take task level means and their errors, then take the mean of these single task means and combine their standard errors hierarchically using Eq.~\eqref{eq:xvar-final}.  For comparison, flattening the same matrix and treating all entries as independent task-level evidence gives:
\begin{equation}
  \widehat{\operatorname{SE}}_{\mathrm{pooled}}
  =\sqrt{\frac{s_X^2}{NM}},
  \qquad
  s_X^2=\frac{1}{NM-1}\sum_{r=1}^{N}\sum_{i=1}^{M}(X_{ri}-\bar X)^2.
\end{equation}
Because averaging is a linear operation and each task has the same number of runs, averaging the task-level means is equivalent to averaging all matrix entries. The pooled SE, however, ignores the hierarchy: additional runs improve estimates of existing tasks rather than providing additional independent tasks. Likewise, simply averaging the individual task SEs does not propagate uncertainty through the second averaging step. At $N=M=20$, both procedures give $\bar X=0.4825$, but the hierarchical SE is $0.07329$ and the pooled SE is $0.02502$, a factor of approximately $2.93$ smaller. The propagated within-task contribution alone has SE $0.01971$. The plot also compares estimates from nested row prefixes, $N\in\{2,5,10,20,50,100\}$, of this same matrix. Thus, the discrepancy is not caused by different observations or a different mean estimator, but by treating repeated executions as additional independent tasks. Panel (d) makes the consequence visible: the pooled SE continues to shrink with more runs, while the hierarchical SE retains uncertainty associated with having evaluated only 20 tasks.
\subsection{\texorpdfstring{$pass@k$}{pass@k} and its Uncertainty} \label{ch:pass-k-uncertainty}

We now consider another important metric in agent evaluation~\cite{chen2021}: $pass@k$, the probability that at least one of $k$ independent attempts succeeds on the same task. The gap between $pass@1$ and $pass@k$ reveals solutions that the current policy can produce but does not produce consistently. This can motivate reinforcement fine-tuning (RFT) aimed at increasing single-attempt success by making successful behaviors more likely~\cite{matsutani2025, lambert2025}. However, $pass@k$ describes repeated sampling from the current policy; it neither predicts nor bounds the performance achievable after training. Given $C_i$ successes in $N$ independent attempts on task $i$, the unbiased estimator of $pass@k$ is
\begin{equation} \label{eq:pass-at-k}
  h_k(C_i) = 1 - \frac{\binom{N-C_i}{k}}{\binom{N}{k}},
  \qquad 1\leq k\leq N.
\end{equation}
Here, $\binom{N-C_i}{k}=0$ when $N-C_i<k$. The estimator averages the at-least-one-success indicator over all size-$k$ subsets drawn without replacement from the observed runs. This subset construction is distinct from the $k$ future independent attempts whose success probability it estimates. Over repeated evaluations, $\mathbb{E}[h_k(C_i)\mid\rho_i]=1-(1-\rho_i)^k$. Before evaluating its standard error, we summarize a few properties of the estimator; a constructive derivation is provided in Appendix~\ref{ch:pass-k-cumulants}:
\begin{itemize}[leftmargin=*]
  \item If $C_i=0$, then $h_k(C_i)=0$ for every admissible $k$: no observed subset contains a success. This does not imply that the true success probability is zero.
  \item If $k=1$, the estimator reduces to the empirical pass rate, $h_1(C_i)=C_i/N$.
  \item For large $N$ with $k$ fixed, the estimator can be approximated as (see also Appendix~\ref{ch:simple-identities}):
    \begin{equation}
      h_k(C_i) \simeq 1 - (1 - p_i)^k,
    \end{equation}
    where $p_i=C_i/N$ is the empirical pass rate for instance $i$. The expression on the right is a plug-in point estimate, not the exact unbiased estimator at finite $N$.

\end{itemize}
Under $C_i\mid\rho_i\sim\operatorname{Binomial}(N,\rho_i)$, the estimator's standard error is the square root of its exact conditional sampling variance:
\begin{equation} \label{eq:var-hk-rho}
  \operatorname{Var}\!\left(h_k(C_i)\mid\rho_i\right)=
  \sum_{c=0}^{N} h_k(c)^2 \binom{N}{c} \rho_i^c(1-\rho_i)^{N-c}-\left(1-(1-\rho_i)^k\right)^2.
\end{equation}
Related finite-sample analysis of the $pass@k$ estimator is presented in the recent manuscript~\cite{anonymous2026passk}. Appendix~\ref{sec:sampling-variance} gives our derivation of these moments using the logarithm of the estimator's characteristic function. The first two cumulants give the mean and variance exactly, without a truncation approximation; further derivatives give higher-order cumulants.

The variance formula is exact at the true probability $\rho_i$. In applications, substituting $p_i=C_i/N$ and taking the square root gives a plug-in estimate of the standard error. This distinction applies to both metrics considered here. In particular, observing $C_i=0$ or $C_i=N$ makes this estimated SE zero, but does not establish that the true probability is exactly zero or one.
\nbfigure[\textwidth][!tb]{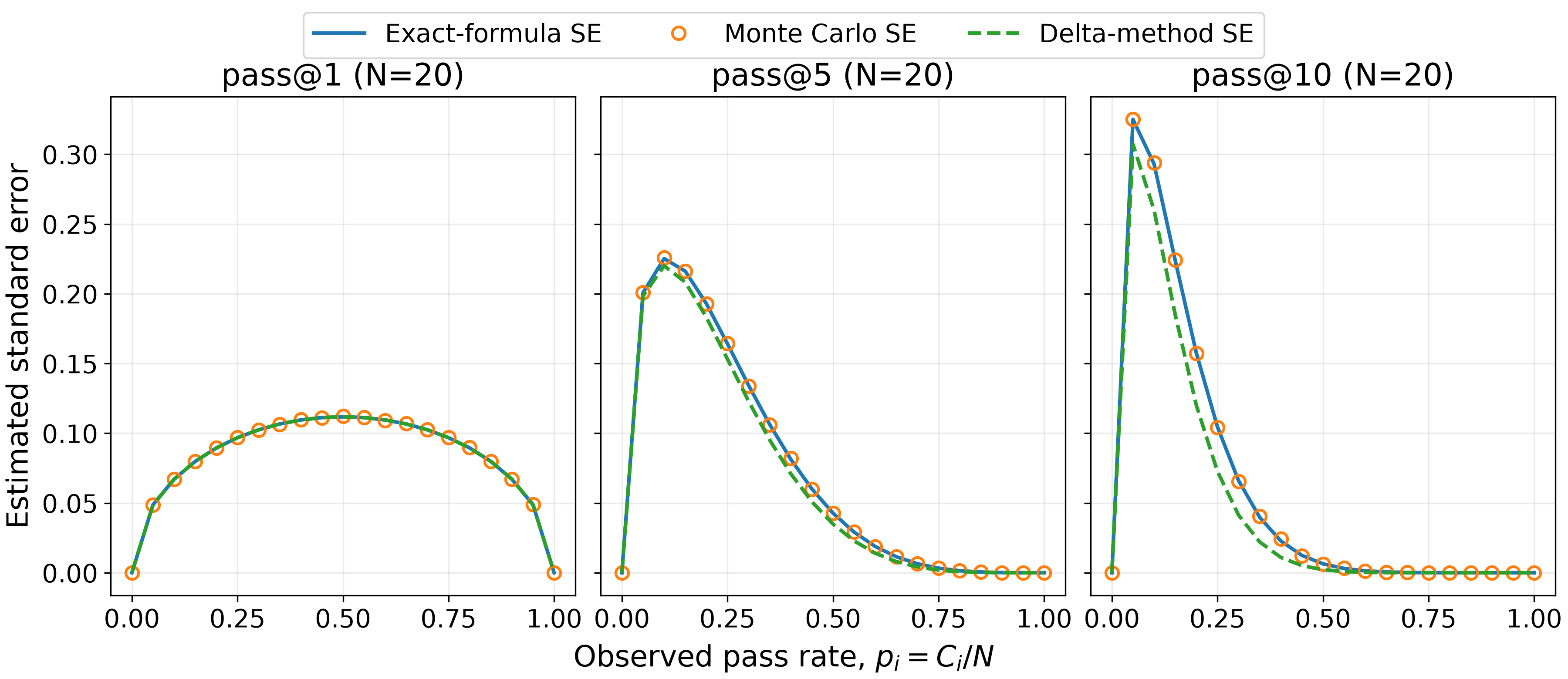}{
  Estimated standard error for $k=1,5,10$ at fixed $N=20$. Solid lines use the exact variance in Eq.~\eqref{eq:var-hk-rho} evaluated at $p_i=C_i/N$; open circles show Monte Carlo estimates under the same fitted binomial model. Dashed lines show the delta-method approximation.
}{se_at_k}
\subsection{\texorpdfstring{$\passhatk$}{pass\textasciicircum k} and its Uncertainty}
Unlike $pass@k$, $\passhatk$, introduced in $\tau$-bench~\cite{yao2024}, is the probability that \emph{all} $k$ independent attempts succeed on the same task. For task $i$ with true success probability $\rho_i$,
\begin{equation} \label{eq:pass-hat-k-prob}
    \passhatk(\rho_i)=\Pr(X_1=1,\ldots,X_k=1\mid\rho_i)=\rho_i^k,
\end{equation}
where $X_1,\ldots,X_k$ are conditionally independent Bernoulli outcomes. This is the success--failure counterpart of $pass@k$: at least one success is the complement of all $k$ attempts failing. Given $C_i$ successes in $N$ independent attempts, the unbiased estimator~\cite{yao2024} is
\begin{equation} \label{eq:pass-hat-k-unbiased}
    \widehat Q_k(C_i) = \frac{\binom{C_i}{k}}{\binom{N}{k}}, \qquad 1\leq k\leq N,
\end{equation}
with $\widehat Q_k(C_i)=0$ if $C_i<k$. This estimator averages the all-success indicator over subsets of $k$ distinct observed runs, selected without replacement, and satisfies $\mathbb{E}[\widehat Q_k(C_i)\mid\rho_i]=\rho_i^k$. Its exact conditional variance is
\begin{equation} \label{eq:pass-hat-k-unbiased-var}
\operatorname{Var}(\widehat Q_k(C_i)\mid\rho_i)=
\sum_{r=1}^{k} \frac{\binom{k}{r}\binom{N-k}{k-r}}{\binom{N}{k}} \left(\rho_i^{\,2k-r} - \rho_i^{\,2k}\right).
\end{equation}
The coefficient in the summand is the probability that two independently and uniformly chosen size-$k$ subsets of the $N$ samples overlap in exactly $r$ attempts; impossible overlaps have coefficient zero. Their shared samples induce the covariance. Both unbiased estimators are U-statistics, and this subset-overlap argument specializes classical U-statistic variance theory~\cite{hoeffdling1948,serfling1980}. Recent manuscripts also analyze finite-sample estimation and uncertainty for $\passhatk$~\cite{anonymous2026passk,anonymous2026rankings}; here, these per-task results are incorporated into the hierarchical uncertainty framework. Replacing $\rho_i$ by $1-\rho_i$ in Eq.~\eqref{eq:pass-hat-k-unbiased-var} gives the variance of the unbiased $pass@k$ estimator in Eq.~\eqref{eq:var-hk-rho}: taking one minus the all-failure estimator leaves its variance unchanged. A simpler alternative estimates $\passhatk$ by substituting the empirical pass rate $p_i=C_i/N$ into $\rho_i^k$:
\begin{equation} \label{eq:pass-hat-k-plugin}
  \widetilde Q_k(C_i)=p_i^k=\left(\frac{C_i}{N}\right)^k.
\end{equation}
Unlike Eq.~\eqref{eq:pass-hat-k-unbiased}, this plug-in estimator averages the all-success indicator over ordered $k$-tuples drawn with replacement from the observed runs. Reusing an observed run is not equivalent to obtaining a fresh independent outcome. For $k>1$ and $0<\rho_i<1$, the estimator is upward biased: because raising a probability to the $k$th power is convex, fluctuations in $p_i$ increase the average of $p_i^k$ above $\rho_i^k$. At $k=1$, both estimators reduce to $p_i$.

Appendix~\ref{ch:pass-hat-k-plugin-moments} gives the exact conditional mean and variance of the plug-in estimator. Its standard error measures variation around its own expectation, not its bias relative to $\rho_i^k$; a smaller SE therefore does not by itself imply a more accurate estimator.
\nbfigure[\textwidth][!tb]{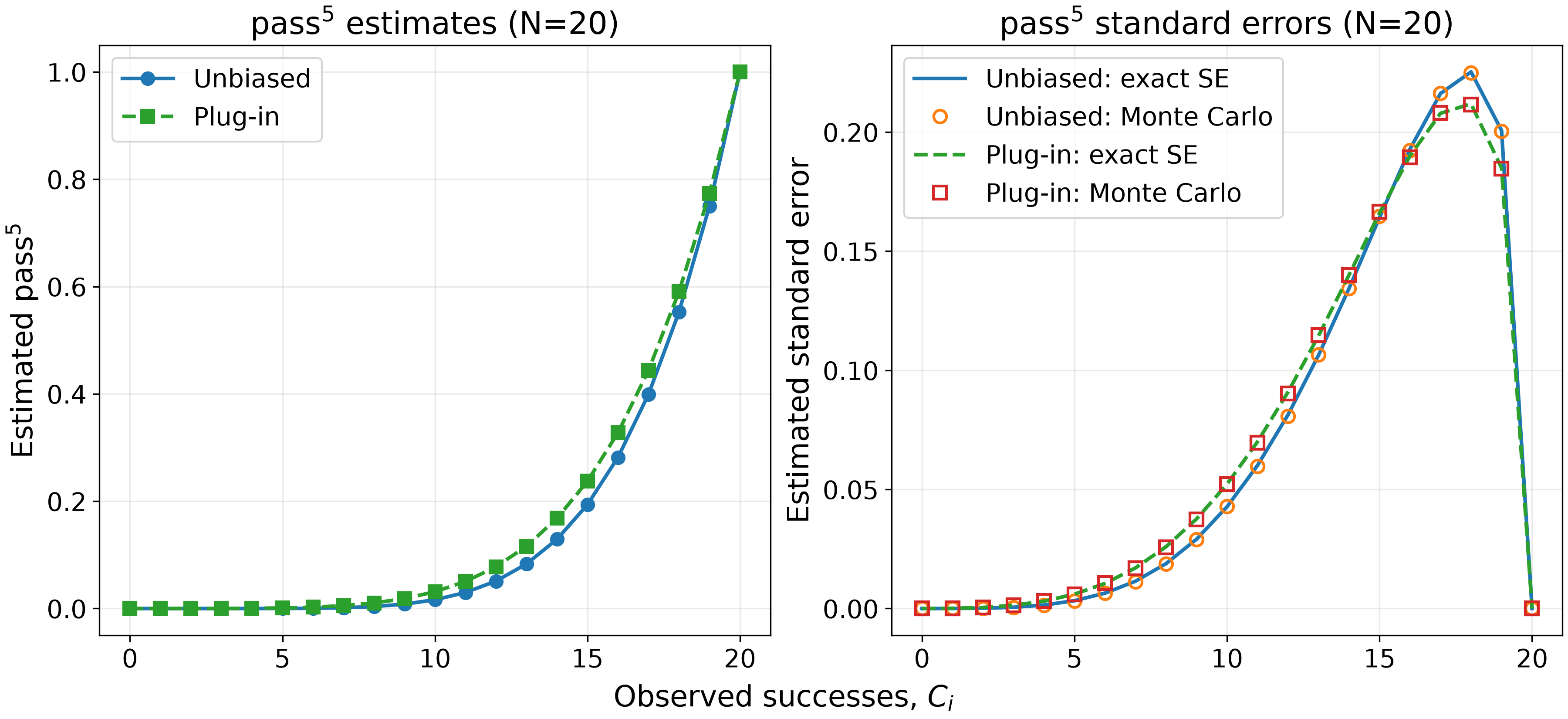}{
  Comparison of the unbiased and plug-in estimators for $\passhatk$ at fixed $N=20$ and $k=5$. (Left) Point estimates as functions of the observed success count $C_i$. (Right) Estimated standard errors from the exact variance formulas evaluated at $p_i=C_i/N$ (lines), compared with Monte Carlo estimates under the same fitted binomial model (open markers).
}{hat_k}
For either estimator, we estimate its standard error by substituting $p_i$ for $\rho_i$ in its exact variance formula and taking the square root. Plugging into a variance formula is a separate choice from using a plug-in point estimator: it can be done for the unbiased estimator as well. Figure~\ref{fig:hat_k} compares the point estimators and their estimated standard errors.

The uncertainty formulas in these two subsections concern individual tasks. Benchmark metrics are obtained by computing the per-task estimates first and then averaging with the chosen task weights; for $k>1$, applying the nonlinear metric to a pooled mean pass rate generally gives a different quantity. For inference about a task population, the law of total variance from Section~\ref{ssec:pass-rate-undertainty} must also account for differences in the underlying per-task metric, $1-(1-\rho_i)^k$ or $\rho_i^k$, across independently sampled tasks. Conditional on a fixed benchmark, only the execution-sampling contribution remains.
\clearpage
\section{Empirical Analysis} \label{sec:empirical-analysis}
In this section we apply the statistical framework developed above to results obtained on the thinkingbox dataset~\cite{thinkingbox}; in particular we apply a the framework introduced in the previous sections to the \emph{Qwen-3.8-27b} model~\cite{qwen38}, via a side-by-side (SBS) analysis of the base and reinforcement fine-tuned model, see Ref.~\cite{thinkingbox} for training details. We provide few additional insights into the comparative performance of the two models, beyond what is reported in the original study. Although we choose to presents details for this specific dataset, the formalism and general analysis prinicple, are independent of the specific benchmark. In the following we refer to the base and reinforcement fine-tuned models' results simply as \emph{base} and \emph{RLFT}, respectively. 

\begin{figure}[!t]
  \centering
  \begin{minipage}[t]{0.49\textwidth}
    \vspace{0pt}
    \includegraphics[width=\linewidth]{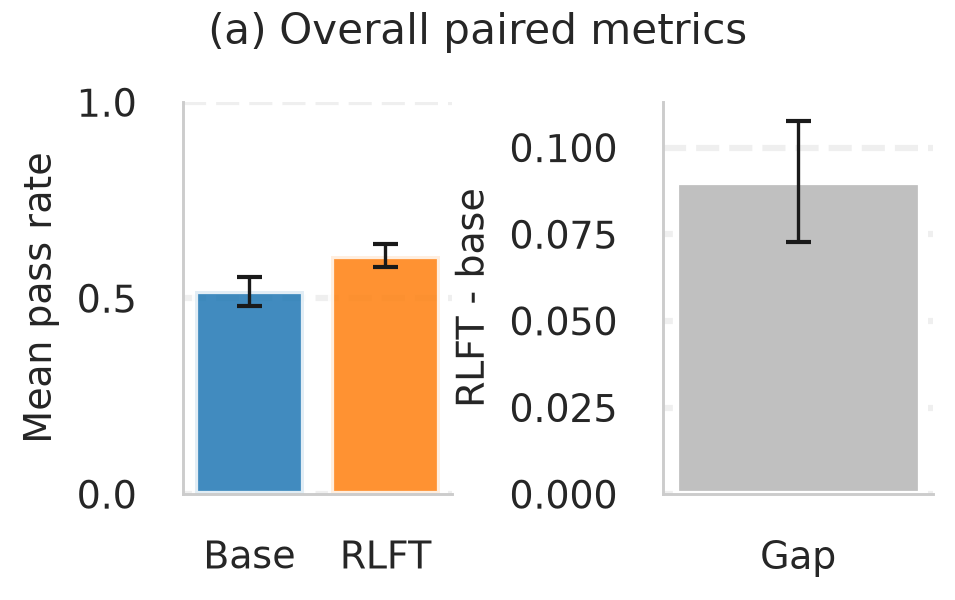}
  \end{minipage}\hfill
  \begin{minipage}[t]{0.49\textwidth}
    \vspace{0pt}
    \includegraphics[width=\linewidth]{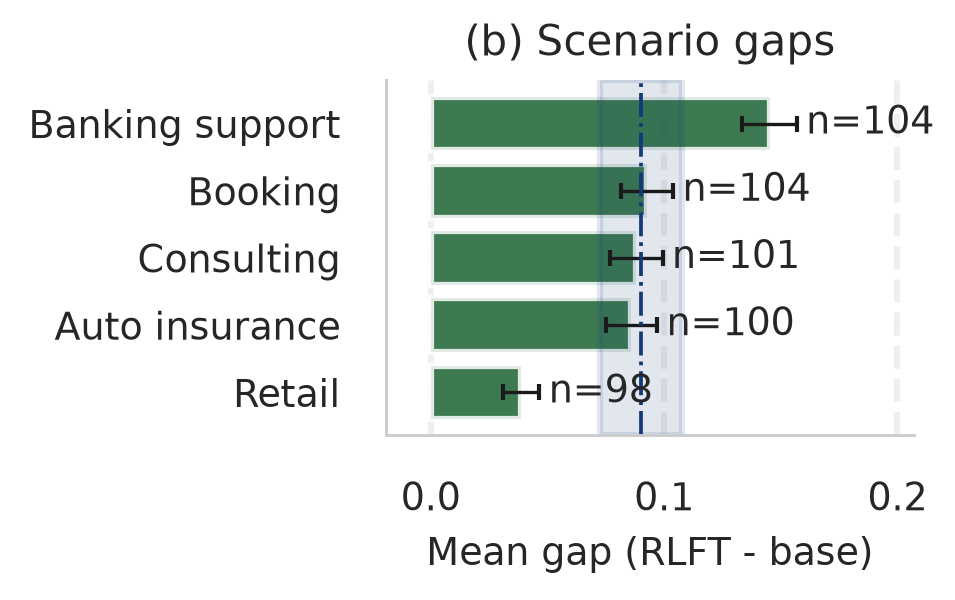}
  \end{minipage}
  \caption{Mean pass-rate comparison between the base and RLFT (candidate) models. Left: overall mean pass rates and their paired gap. Right: scenario-level performance gaps (RLFT minus base), with instance counts $n$. Error bars indicate standard errors; the blue line and shaded band indicate the overall mean gap and one standard error. The scenario aggregated metrics do not reveal the full picture of model performance across different business scenarios.}
  \label{fig:gap-comparison}
\end{figure}

The benchmark consists of five different business scenarios, each comprising $\sim 100$ evaluation instances. Overall metrics aggregate across attempts, tasks, and scenarios using the equal averaging defined above. We first consider mean pass-rate at different level of aggregation, as explained in the previous sections. Fig.~\eqref{fig:gap-comparison} shows the comparison between the base and RLFT models, both overall and at the scenario level. In particular we focus on the performance gap and report it alongside the corresponding standard errors, propagated according to the statistical framework introduced above and again on the gap itself. Perhaps not surprisingly, the magnitude of the gap varies across scenarios, highlighting differering learning dynamics and the varying impact of reinforcement fine-tuning across different business contexts. The disparity itself depends on the base model (the prior), but can be further increased or mitigated by the fine-tuning process. 
We can also zoom in at the instance level to better understand the gap distribution and the error characteristics induced by the training dynamics. Fig.~\eqref{fig:gap_neobank} provides an example of such an instance-level analysis, highlighting how individual evaluation instances contribute to the overall performance gap and its variability.
\nbfigure[\textwidth][!t]{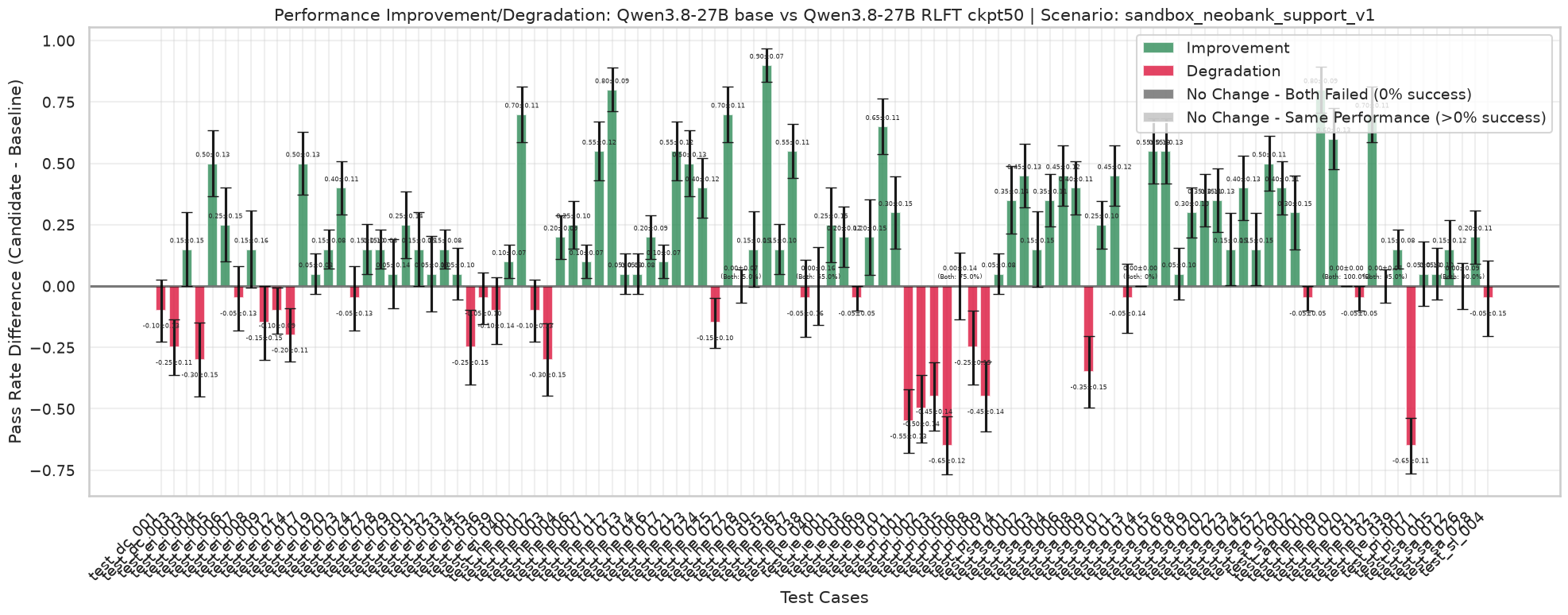}{
  Distribution of the performance gap across individual instances in the $104$-instance banking support scenario, highlighting variability and error characteristics. Green indicates improvements while red indicates regressions. No change is shown in gray.
}{gap_neobank}
Comparing to the scenario-level analysis in Fig.~\ref{fig:gap-comparison}, the instance-level view reveals the heterogeneity in model performance across individual evaluation instances, and also how individual errors are washed out in the aggregated metrics, that is the essence of coarse-graining, see Section~\ref{sec:statistical-analysis-framework}. Figure~\ref{fig:gap_neobank} shows only the 104 instances in the banking support scenario, comparing the same instance across the two models. Table~\ref{tab:scenario-comparison} reports counts separately for each scenario, including instances with increased, decreased, or unchanged observed pass rates. Within each scenario, we count changes whose per-instance approximate $95\%$ Wald intervals exclude zero: the interval is the observed RLFT-minus-base pass-rate difference plus or minus $1.96$ times its estimated standard error, obtained by adding the two binomial plug-in variances under conditional independence of the models' executions.
\begin{table}[!htbp]
  \centering
  \small
  \resizebox{\textwidth}{!}{%
    \begin{tabular}{@{}lccccc@{}}
      \toprule
      Metric
      & \shortstack{booking}
      & \shortstack{auto insurance}
      & \shortstack{consulting services}
      & \shortstack{retail}
      & \shortstack{banking support} \\
      \midrule
      number of instances & 104 & 100 & 101 & 98 & 104 \\
      baseline mps $\pm$ ste & $0.534 \pm 0.033$ & $0.479 \pm 0.032$ & $0.457 \pm 0.035$ & $0.640 \pm 0.041$ & $0.479 \pm 0.030$ \\
      candidate mps $\pm$ ste & $0.627 \pm 0.035$ & $0.565 \pm 0.040$ & $0.545 \pm 0.037$ & $0.679 \pm 0.042$ & $0.624 \pm 0.034$ \\
      mean gap $\pm$ ste & $0.093 \pm 0.011$ & $0.086 \pm 0.011$ & $0.088 \pm 0.011$ & $0.039 \pm 0.008$ & $0.145 \pm 0.012$ \\
      \midrule
      tests improved & 62 & 58 & 57 & 45 & 69 \\
      tests degraded & 24 & 28 & 30 & 16 & 28 \\
      tests no change & 18 & 14 & 14 & 37 & 7 \\
      significant changes (95\% CI) & 35 & 32 & 41 & 10 & 46 \\
      no change total & 18 & 14 & 14 & 37 & 7 \\
      no change both zero & 6 & 10 & 8 & 10 & 1 \\
      no change both same nonzero & 12 & 4 & 6 & 27 & 6 \\
      \bottomrule
    \end{tabular}%
  }
  \caption{Separate comparisons of the baseline and candidate models within each scenario. Each column reports scenario-level mean pass rates (mps), their standard errors and candidate-minus-baseline gaps, together with counts of instance-level changes. No aggregation across scenarios is performed in this table.}
  \label{tab:scenario-comparison}
\end{table}
\subsection{Task Coverage, Reliability and Distributional Shifts}
\begin{figure}[t]
  \centering
  \begin{minipage}[t]{0.4\textwidth}
    \vspace{0pt}
    \includegraphics[width=\linewidth]{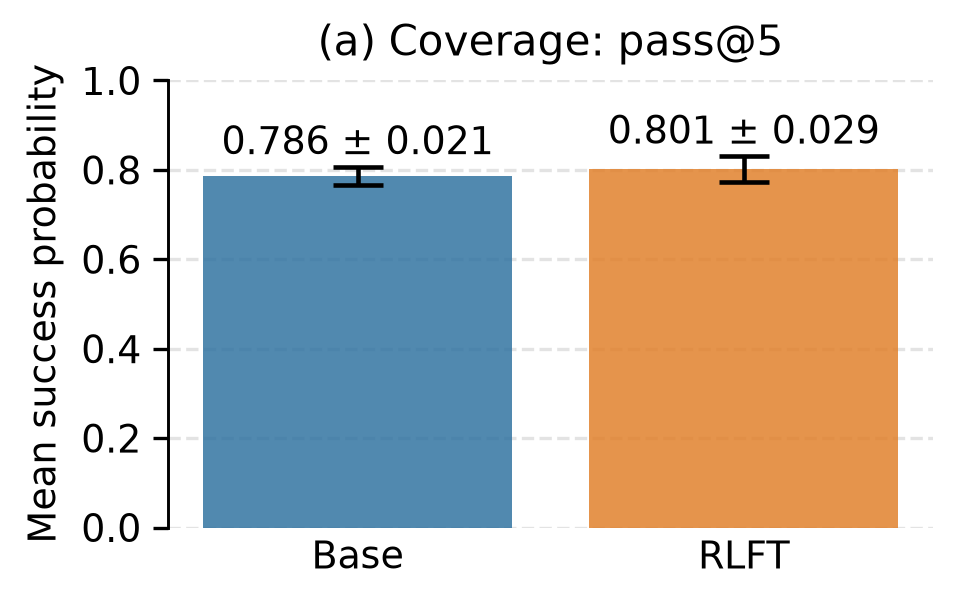}
  \end{minipage}\hspace{0.02\textwidth}%
  \begin{minipage}[t]{0.4\textwidth}
    \vspace{0pt}
    \includegraphics[width=\linewidth]{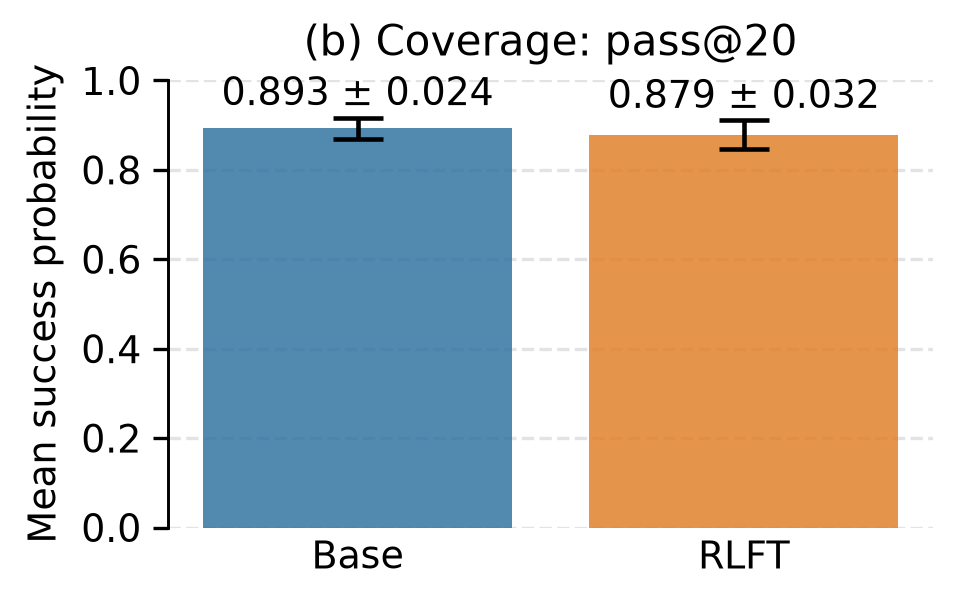}
  \end{minipage}

  \smallskip
  \begin{minipage}[t]{0.4\textwidth}
    \vspace{0pt}
    \includegraphics[width=\linewidth]{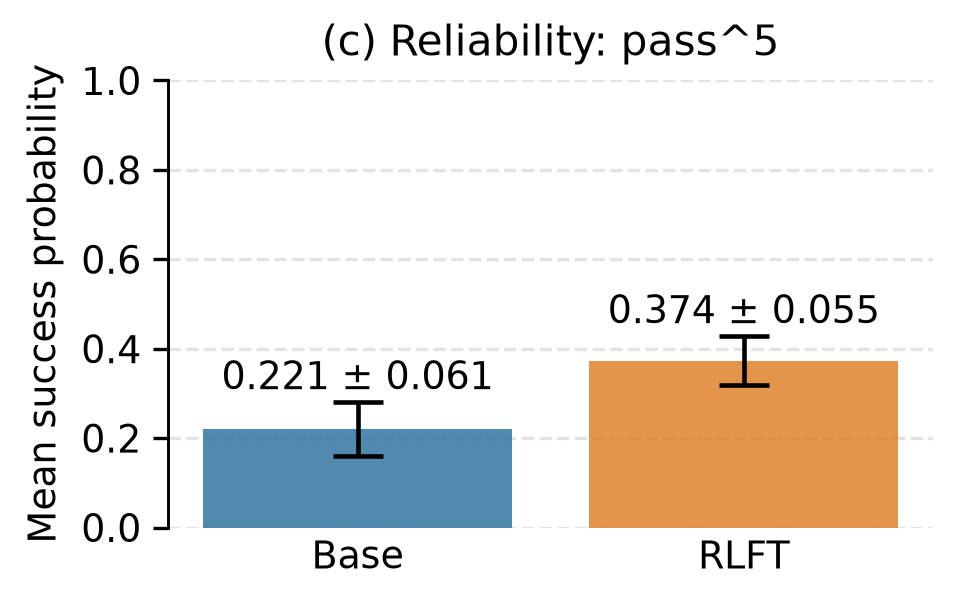}
  \end{minipage}\hspace{0.02\textwidth}%
  \begin{minipage}[t]{0.4\textwidth}
    \vspace{0pt}
    \includegraphics[width=\linewidth]{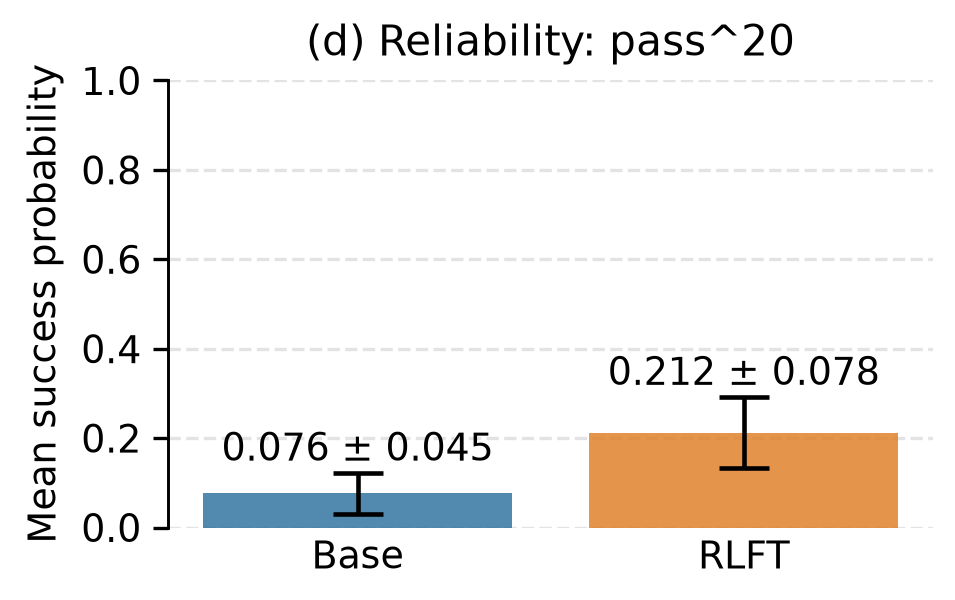}
  \end{minipage}
  \caption{Blue: Base model, Orange: Candidate model. Top row: mean $pass@k$ (task coverage). Bottom row: mean $\passhatk$ (reliability). Left column: $k=5$; right column: $k=20$. Error bars indicate propagated standard errors.}
  \label{fig:pass_at_k}
\end{figure}

While $Pass@1$ (pass-rate) measures average success probability, larger $k$ increasingly measures coverage: whether a task has any appreciable chance of being solved. Fig~\eqref{fig:pass_at_k} shows the mean pass@5 and pass@20 across scenarios for the baseline and candidate models. The two plots tell a differnt story: while a significant improvement is observed at $pass@1$, already at $k=5$ the difference between the two models is not statistically significant, and at $k=20$ the performance is almost indistinguishable. For the overall task-population comparison, the approximate $95\%$ confidence intervals for the mean differences at $k=5$ and $k=20$ both include zero. This pattern is consistent with the interpretation that the RL fine-tuned model mainly improves the probability of success for tasks that are already solvable by the baseline, without substantially increasing coverage for harder tasks; in the next section we show the same effect by looking at the distributional shift across the two models.
%
Although the task-coverage point estimates change little in our candidate model, reliability shows a more noticeable improvement, suggesting that the candidate model is more consistent in solving tasks it can handle, even if it does not cover substantially more tasks overall. Although the two metrics are related, they capture different aspects of model performance, and improvements in reliability can be valuable even when task coverage remains similar.

Finally, we examine the effect of model training within a distributional shift point of view. Although this analysis is not part of our discussion on uncertaining quantification, it nevertheless provides a more holistic representation of the effect of model training on the task performance. Fig.~\eqref{fig:rl_distr} illustrates the pass rate distribution before and after fine tuning. Two clear trends can be observed: first, there is a noticeable shift towards higher pass rates for the candidate model, particularly in the high pass-rate region; second, the distribution indicates that the model is becoming more reliable in the Goldilocks Zone, where task difficulty is optimal for learning. Finally, some of the tasks that had low pass rate, and were previously outside of the Goldilocks Zone, moved into the Goldilocks Zone and experienced improved success rates, while few regressed. Understading regression behaviour is paramount for evaluating model improvements and to develop novel training strategies to mitigate negative impacts on previously well-performing tasks.
\nbfigure[0.8\textwidth][!t]{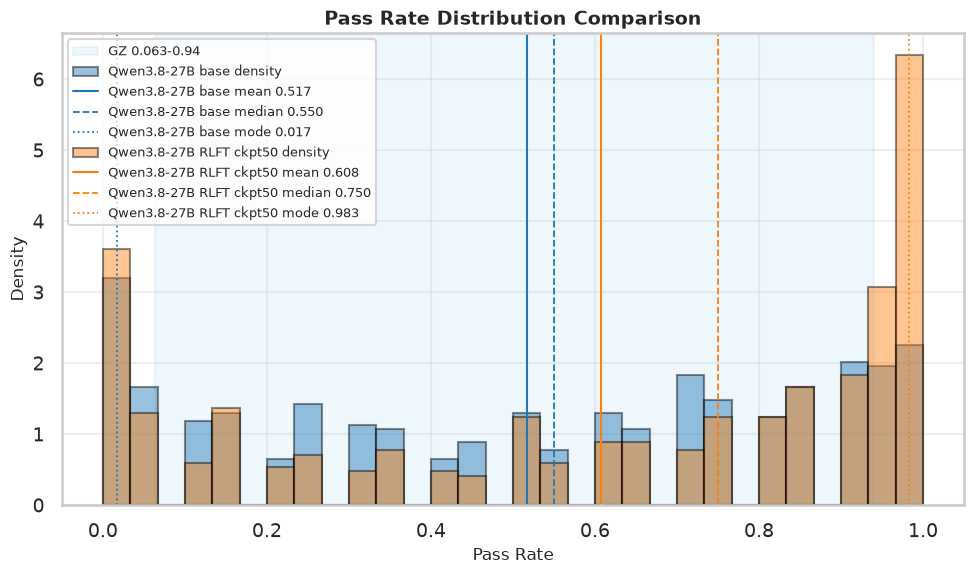}{
  Distributional shift of model performance across different scenarios of mean pass rate for the baseline and candidate models across different bins of task difficulty. Goldilocks Zone indicates the range of task difficulty where the moddel can learn more efficiently. The figures identifies a push towards higher pass rates for the candidate model, particuarly in the high pass-rate region.
}{rl_distr}
\section{Conclusion} \label{sec:conclusion}
Reliable evaluation of agentic systems requires more than an aggregate success score: it requires identifying what a metric measures, how precisely it has been estimated, and which differences are hidden by aggregation. We have presented a statistical framework organized around scenarios, task instances, and repeated samples, distinguishing stochastic execution variability from finite-sample estimation uncertainty and heterogeneity in task difficulty. Connecting this hierarchy to coarse-graining makes explicit both the value and the limitations of benchmark averages: they summarize collective behavior, but cannot by themselves characterize instance-level reliability or reveal localized regressions.

Within this framework, we have analyzed mean pass rate, $pass@k$, and $\passhatk$, including exact finite-sample sampling variances for the unbiased estimators of the latter two metrics and hierarchical uncertainty propagation. These results give a principled basis for attaching uncertainty estimates to commonly reported scores and interpreting model comparisons. They also clarify that single-attempt performance, task coverage under retries, and repeated-success reliability answer different questions; no one of these metrics is a substitute for the others.

The Thinkingbox case study~\cite{thinkingbox} illustrates the practical importance of this distinction. For the base and reinforcement fine-tuned Qwen-3.8-27b models considered here, improved mean pass rate coexists with similar coverage at larger $k$ and higher point estimates of repeated-success reliability. Together with the instance-level gaps and the shift toward higher pass rates, this pattern shows that the fine-tuned model is making already-solvable tasks more consistently successful, rather than substantially expanding task coverage, although some previously unsolved tasks may still benefit from fine-tuning, as shown by the instance-level and distributional analyses. Scenario averages nevertheless conceal individual improvements and regressions, emphasizing why aggregate gains should be examined alongside paired task-level comparisons and their uncertainty. 
 
The resulting recommendation is to report mean pass rate, coverage, and reliability jointly, with uncertainty estimates at the relevant aggregation levels and sufficient instance-level detail to expose regressions. Extending this analysis to dependent executions and evaluation-budget allocation is a natural direction for future work. Ultimately, progress in agentic systems should be judged not only by whether average performance increases, but by whether that improvement is statistically supported, broadly distributed across tasks, and aligned with the reliability users actually need.

\section*{Use of Large Language Models (LLMs)}
\label{sec:appendix_llm}
We used large language models (LLMs) to assist with manuscript drafting and editing, as well as code and math development. The authors developed the research ideas and methodology, derived main mathematical expressions, conducted the experiments, analyzed the results, and reviewed the LLM-assisted content and code. The authors take full responsibility for the final content of this work.

\bibliographystyle{unsrt}
\bibliography{agentstats}

@article{thinkingbox,
  author  = {Zhuochun Li and Youngmin Ko and Ali Keramati and Tuhin Kundu and Liang-Chun Tsai and Nicola Ferri and Mirco Milletari\' and Jiaxiang Liu and Susana Palmaz Lopez Pelaez and Yuepeng Wang and Vadim Smolyakov and Xiang Jiang and Kjartan \'{O}lafsson and Tommy Guy},
  title   = {One Success Isn't Reliability: Thinkingbox, a Sandbox and Benchmark for Agents in Stateful Business Workflows},
  journal = {arXiv preprint arXiv:2608.19741},
  year    = {2026}
}

@article{chen2021,
  title   = {Evaluating Large Language Models Trained on Code},
  author  = {Mark Chen and Jerry Tworek and Heewoo Jun and Qiming Yuan and Henrique Ponde de Oliveira Pinto and Jared Kaplan and Harri Edwards and Yuri Burda and Nicholas Joseph and Greg Brockman and Alex Ray and Raul Puri and Gretchen Krueger and Michael Petrov and Heidy Khlaaf and Girish Sastry and Pamela Mishkin and Brooke Chan and Scott Gray and Nick Ryder and Mikhail Pavlov and Alethea Power and Lukasz Kaiser and Mohammad Bavarian and Clemens Winter and Philippe Tillet and Felipe Petroski Such and Dave Cummings and Matthias Plappert and Fotios Chantzis and Elizabeth Barnes and Ariel Herbert-Voss and William Hebgen Guss and Alex Nichol and Alex Paino and Nikolas Tezak and Jie Tang and Igor Babuschkin and Suchir Balaji and Shantanu Jain and William Saunders and Christopher Hesse and Andrew N. Carr and Jan Leike and Josh Achiam and Vedant Misra and Evan Morikawa and Alec Radford and Matthew Knight and Miles Brundage and Mira Murati and Katie Mayer and Peter Welinder and Bob McGrew and Dario Amodei and Sam McCandlish and Ilya Sutskever and Wojciech Zaremba},
  journal = {arXiv preprint arXiv:2107.03374},
  year    = {2021}
}

@inproceedings{huang-etal-2025-crmarena,
    title = "CRMArena: Understanding the Capacity of LLM Agents to Perform Professional CRM Tasks in Realistic Environments",
    author = "Huang, Kung-Hsiang  and
      Prabhakar, Akshara  and
      Dhawan, Sidharth  and
      Mao, Yixin  and
      Wang, Huan  and
      Savarese, Silvio  and
      Xiong, Caiming  and
      Laban, Philippe  and
      Wu, Chien-Sheng",
    booktitle = "Proceedings of the 2025 Conference of the Nations of the Americas Chapter of the Association for Computational Linguistics: Human Language Technologies (Volume 1: Long Papers)",
    year = "2025",
}

@article{huang-etal-2026-crmarena-pro,
    title = "CRMArena-Pro: Holistic Assessment of LLM Agents Across Diverse Business Scenarios and Interactions",
    author = "Huang, Kung-Hsiang  and
      Prabhakar, Akshara  and
      Thorat, Onkar  and
      Agarwal, Divyansh  and
      Choubey, Prafulla Kumar  and
      Mao, Yixin  and
      Savarese, Silvio  and
      Xiong, Caiming  and
      Wu, Chien-Sheng",
    journal = "Transactions on Machine Learning Research (TMLR)",
    year = "2026",
}

@misc{bandi2026mcpatlas,
  title         = {MCP-Atlas: A Large-Scale Benchmark for Tool-Use Competency with Real MCP Servers},
  author        = {Bandi, Chaithanya and Dumitru, Razvan-Gabriel and Hertzberg, Ben and Agarwal, Divyansh and Boo, Geobio and Polakam, Tejas and Hassaan, Sami and Da, Jeff and Kim, HiJae and Gupta, Vipul and Sharma, Manasi and Park, Andrew and Dimakis, Martin and Hernandez Montoya, Ernesto Gabriel and Rambado, Dan and Salazar, Ivan and Cruz, Rafael and Rezaei, MohammadHossein and Rane, Chetan and Levin, Ben and Zhang, Daniel Yue and Kenstler, Brad and Liu, Bing},
  year          = {2026},
  eprint        = {2602.00933},
  archivePrefix = {arXiv},
  primaryClass  = {cs.SE},
  url           = {https://arxiv.org/abs/2602.00933}
}

@misc{anonymous2026passk,
  title  = {Unbiased pass@k and pass{\textasciicircum}k evaluation with fewer rollouts},
  author = {Anonymous},
  year   = {2026},
  note   = {Submitted to ICLR 2027. OpenReview: https://openreview.net/forum?id=SWs3hMG9IC}
}

@misc{anonymous2026rankings,
  author = {Anonymous},
  title = {Are Agent Reliability Rankings Real? A Statistical Treatment of the pass{\textasciicircum}k Metric},
  year = {2026},
  note   = {Submitted to ICLR 2027. OpenReview: https://openreview.net/forum?id=NH98Pcbbwd}
}

@article{miller2024,
  author  = {Evan Miller},
  title   = {Adding Error Bars to Evals: A Statistical Approach to Language Model Evaluations},
  journal = {arXiv preprint arXiv:2411.00640},
  year    = {2024}
}

@article{rabanser2026reliability,
  author  = {Stephan Rabanser and Sayash Kapoor and Peter Kirgis and Kangheng Liu and Saiteja Utpala and Arvind Narayanan},
  title   = {Towards a Science of {AI} Agent Reliability},
  journal = {arXiv preprint arXiv:2602.16666},
  year    = {2026},
  url     = {https://arxiv.org/abs/2602.16666v2}
}

@article{yao2024,
  author  = {Shunyu Yao and Noah Shinn and Pedram Razavi and Karthik Narasimhan},
  title   = {$\tau$-bench: A Benchmark for Tool-Agent-User Interaction in Real-World Domains},
  journal = {arXiv preprint arXiv:2406.12045},
  year    = {2024},
  url     = {https://arxiv.org/abs/2406.12045}
}

@inproceedings{hariri2026,
  author    = {Mohsen Hariri and Amirhossein Samandar and Michael Hinczewski and Vipin Chaudhary},
  title     = {Don't {Pass@k}: A {Bayesian} Framework for Large Language Model Evaluation},
  booktitle = {The Fourteenth International Conference on Learning Representations},
  year      = {2026},
  url       = {https://arxiv.org/abs/2510.04265}
}

@article{bjarnason2026,
  author  = {Bjarni Haukur Bjarnason and Andr{\'e} Silva and Martin Monperrus},
  title   = {On Randomness in Agentic Evals},
  journal = {arXiv preprint arXiv:2602.07150},
  year    = {2026},
  url     = {https://arxiv.org/abs/2602.07150}
}

@misc{dewasurendra2026,
  author       = {Pahan Dewasurendra},
  title        = {Extrapolation Is an Assumption: Sharp Limits for {Pass@k} Forecasting},
  howpublished = {\url{https://doi.org/10.5281/zenodo.21698609}},
  year         = {2026}
}

@inproceedings{bowyer2025,
  author    = {Bowyer, Sam and Aitchison, Laurence and Ivanova, Desi R.},
  title     = {Position: Don't Use the CLT in LLM Evals With Fewer Than a Few Hundred Datapoints},
  booktitle = {Proceedings of the $42^{nd}$ International Conference on Machine Learning, Vancouver, Canada, PMLR 267},
  year      = {2025}
}

@book{parisi1,
  author    = {Leuzzi, Luca and Marinari, Enzo and Parisi, Giorgio},
  title     = {Probability Theory for quantitative scientists},
  publisher = {Cambridge University Press},
  year      = {2025}
}

@book{blitzstein2019,
  author    = {Blitzstein, Joseph K. and Hwang, Jessica},
  title     = {Introduction to Probability},
  edition   = {2},
  publisher = {Chapman and Hall/CRC},
  year      = {2019}
}

@book{tao2011,
  author    = {Tao, Terence},
  title     = {An Introduction to Measure Theory},
  series    = {Graduate Studies in Mathematics},
  volume    = {126},
  publisher = {American Mathematical Society},
  year      = {2011},
  isbn      = {978-0-8218-6919-2},
  url       = {https://terrytao.wordpress.com/books/an-introduction-to-measure-theory/}
}

@book{villani2003,
  author    = {Villani, C{\'e}dric},
  title     = {Topics in Optimal Transportation},
  series    = {Graduate Studies in Mathematics},
  volume    = {58},
  publisher = {American Mathematical Society},
  year      = {2003},
  url       = {https://bookstore.ams.org/gsm-58/}
}

@book{santambrogio2015,
  author    = {Santambrogio, Filippo},
  title     = {Optimal Transport for Applied Mathematicians: Calculus of Variations, {PDEs}, and Modeling},
  series    = {Progress in Nonlinear Differential Equations and Their Applications},
  volume    = {87},
  publisher = {Birkh{\"a}user},
  year      = {2015},
  doi       = {10.1007/978-3-319-20828-2}
}

@book{kardar2007,
  author    = {Kardar, Mehran},
  title     = {Statistical Physics of Particles},
  publisher = {Cambridge University Press},
  year      = {2007}
}

@article{matsutani2025,
  author  = {Matsutani, Kohsei and Takashiro, Shota and Minegishi, Gouki and Kojima, Takeshi and Iwasawa, Yusuke and Matsuo, Yutaka},
  journal = {arXiv preprint arXiv:2509.21128},
  title   = {RL Squeezes, SFT Expands: A Comparative Study of Reasoning LLMs},
  year    = {2025}
}

@misc{lambert2025,
  author       = {N. Lambert},
  title        = {Reinforcement Learning from Human Feedback},
  howpublished = {\url{https://rlhfbook.com/}},
  year         = {2025}
}

@misc{variance-wiki,
  author       = {Wikipedia contributors},
  title        = {Law of Total Variance},
  howpublished = {\url{https://en.wikipedia.org/wiki/Law_of_total_variance}},
  note         = {Wikipedia}
}

@article{hoeffdling1948,
  title   = {A Class of Statistics with Asymptotically Normal Distribution},
  author  = {Hoeffding, Wassily},
  journal = {The Annals of Mathematical Statistics},
  volume  = {19},
  number  = {3},
  pages   = {293--325},
  year    = {1948},
  doi     = {10.1214/aoms/1177730196}
}

@book{serfling1980,
  author    = {Serfling, Robert J.},
  title     = {Approximation Theorems of Mathematical Statistics},
  publisher = {Wiley},
  year      = {1980}
}

@misc{qwen38,
    title = {{Qwen3.8-27B} Model Card},
    howpublished = {\url{https://huggingface.co/Qwen/Qwen3.8-27B}},
    url = {https://huggingface.co/Qwen/Qwen3.8-27B},
    author = {{Qwen Team}},
    year = {2026}
}


\appendix
\numberwithin{equation}{section}

\section{Law of total variance: a coarse-graining derivation} \label{ch:metrics-coarse-graining}

In this section, we provide an explicit evaluation of the hierarchical model described in Section~\ref{sec:statistical-analysis-framework} using coarse graining methods. We use coarse-graining in the sense of replacing a detailed description by distributions of aggregate observables, a standard
viewpoint in statistical mechanics. Here, the corresponding probabilistic operations are aggregation and marginalization, and the resulting variance decomposition is the law of total variance~\cite{blitzstein2019}. We evaluate these operations explicitly using the methods of Refs.~\cite{parisi1,kardar2007}.

We consider one scenario with $M$ instances and $N$ runs per instance. As in Section~\ref{sec:statistical-analysis-framework}, we suppress the scenario index $s$ and write $\xi_{ir}$ for the binary outcome of run $r$ on instance $i$. For simplicity, each instance is sampled the same number of times; the derivation extends directly to variable sample counts. For each instance $i$, define the per-instance average over runs:
\begin{equation} \label{eq:coarse-grain-1}
  \bar{\xi}_i = \frac{1}{N} \sum_{r=1}^{N} \xi_{ir} = \frac{C_i}{N},
\end{equation}
where $C_i$ counts the successful runs for instance $i$. Conditional on its true success probability $\rho_i$, each outcome is Bernoulli:
\begin{equation} \label{eq:bernoulli-likelihood}
  \mathbb{P}(\xi_{ir} \mid \rho_i) = \rho_i \, \delta(\xi_{ir} - 1) + (1 - \rho_i) \, \delta(\xi_{ir}).
\end{equation}
We assume the $\rho_i$ are i.i.d. draws from a distribution $g(\rho)$, representing the variability in instance difficulty. The scenario-level average is
\begin{equation} \label{eq:coarse-grain-2}
  \hat{\xi} = \frac{1}{M} \sum_{i=1}^{M} \bar{\xi}_i.
\end{equation}
For binary outcomes, $\bar{\xi}_i$ is the instance pass rate $p_i$ from Section~\ref{sec:statistical-analysis-framework}, and $\hat{\xi}$ is the scenario pass rate $\bar p$. The calculation proceeds in two steps: first, we evaluate the conditional probability distribution of $\bar{\xi}_i$ given $\rho_i$; then, we marginalize over the instance probabilities to obtain the unconditional distribution of $\hat{\xi}$. Following Ref.~\cite{parisi1}, the conditional distribution can be evaluated as follows:
\begin{align}
  \mathbb{P}_N(\bar \xi_i \mid \rho_i) &= \left\langle\delta\!\left(\bar{\xi}_i - \frac{1}{N} \sum_{r=1}^{N} \xi_{ir}\right) \right\rangle_{\xi_{ir} \mid \rho_i} =\int \prod_{r=1}^N d \xi_{ir}\, \mathbb{P}(\xi_{ir} \mid \rho_i)\delta\!\left(\bar{\xi}_i-\frac{1}{N}\sum_{r=1}^N \xi_{ir}\right), \nonumber \\
  &= \int \frac{d \lambda }{2\pi} e^{i \lambda \bar{\xi}_i} \prod_{r=1}^N \int d \xi_{ir} \mathbb{P}(\xi_{ir} \mid \rho_i) e^{-i \frac{\lambda}{N} \xi_{ir}} = \int \frac{d \lambda }{2\pi} e^{i \lambda \bar{\xi}_i} \left[\rho_i e^{-i \frac{\lambda}{N}} + (1 - \rho_i)\right]^N, \nonumber \\
  &= \int \frac{d \lambda }{2\pi} e^{i \lambda \bar{\xi}_i} \chi_i^N(\lambda/N \mid \rho_i) \label{eq:conditional_1}
\end{align}
where we have used the Fourier representation of the Dirac delta function
\begin{equation}
  \delta(x) = \int \frac{d \lambda}{2\pi} e^{i \lambda x},
\end{equation}
with $\lambda$ a Fourier variable enforcing the Dirac delta constraint. The one-run characteristic function is $\chi_i(t\mid\rho_i)=\mathbb{E}[e^{-it\xi_{ir}}\mid\rho_i]=(1-\rho_i)+\rho_i e^{-it}$; independence of the $N$ runs gives the $N$th power in Eq.~\eqref{eq:conditional_1}. From here, we could evaluate the conditional moments of $\bar \xi_i$ given $\rho_i$ by taking derivatives of its characteristic function. We move instead to the second level of the hierarchy and first evaluate the conditional distribution of $\hat \xi$ given the set $\rho_{1:M}$:
\begin{align}
  \mathbb{P}_{M,N}(\hat \xi \mid \rho_{1:M})
  &= \int \prod_{i=1}^M d\bar{\xi}_{i}\,\mathbb{P}_N(\bar{\xi}_{i} \mid \rho_i)
  \,\delta\!\left(\hat{\xi}-\frac{1}{M}\sum_{i=1}^M \bar{\xi}_{i}\right) \nonumber \\
  &= \int \frac{d\lambda}{2\pi} e^{i\lambda\hat{\xi}}
  \,\prod_{i=1}^M \chi_i^N\!\left(\frac{\lambda}{MN} \mid \rho_i\right) \nonumber \\
  &= \int \frac{d\lambda}{2\pi}
  \,\exp\!\left(i\lambda\hat{\xi}
  +\sum_{i=1}^M N\Phi_i\!\left(\frac{\lambda}{MN} \mid \rho_i\right)\right) \label{eq:conditional_2}
\end{align}
The conditional characteristic function appearing in the inverse Fourier transform in Eq.~\eqref{eq:conditional_2} is $\chi_{\hat{\xi}}(\lambda \mid \rho_{1:M})=\prod_{i=1}^M\chi_i^N(\lambda/(MN)\mid\rho_i)$. We define the cumulant-generating function for the one-run characteristic function by $\Phi_i(t \mid \rho_i)=\log\chi_i(t \mid \rho_i)$, where $t=\lambda/(MN)$. For large $M,N$ we expand around $t_0=0$, writing $t=t_0+\delta t$, and retain terms through second order in $\delta t$:
\begin{equation} \label{eq:cumulant_expansion}
  \Phi_i(t \mid \rho_i) = -i\rho_i \delta t - \frac{1}{2}\rho_i(1-\rho_i) \delta t^2 + O(\delta t^3).
\end{equation}
Substituting this expansion into the logarithm of the conditional characteristic function in Eq.~\eqref{eq:conditional_2} gives, through second order in $\lambda$,
\begin{equation} \label{eq:conditional_3}
  \log\chi_{\hat{\xi}}(\lambda \mid \rho_{1:M})
  = -\frac{i\lambda}{M}\sum_{i=1}^M \rho_i
  - \frac{\lambda^2}{2M^2N}\sum_{i=1}^M \rho_i(1-\rho_i) + O(\lambda^3).
\end{equation}
This expansion identifies the first two conditional cumulants exactly; no normal approximation is needed to obtain the moments. Thus, conditional on $\rho_{1:M}$,
\begin{align} \label{eq:conditional_moments}
  \mathbb{E}[\hat \xi \mid \rho_{1:M}] &= \frac{1}{M} \sum_{i=1}^M \rho_i, \\
  \operatorname{Var}(\hat \xi \mid \rho_{1:M}) &= \frac{1}{M^2 N} \sum_{i=1}^M \rho_i(1-\rho_i).
\end{align}

\subsection{Marginal Characteristic function}

We can now marginalize the conditional characteristic function over the instance probabilities to obtain the unconditional moments of $\hat \xi$. Under the i.i.d. assumption, this gives:
\begin{align} \label{eq:marginalization}
  \chi_{\hat{\xi}}(\lambda)
  &= \int_0^1 \left[\prod_{i=1}^{M} g(\rho_i)\,d\rho_i\right]
  \chi_{\hat{\xi}}(\lambda \mid \rho_{1:M})
\end{align}
Assuming that instances are i.i.d. draws from $g(\rho)$, we can use a moment expansion to second order similar to what done in Eq.~\ref{eq:cumulant_expansion}. This only assumes that the first two moments of $g(\rho)$ are finite, but otherwise makes no assumption on the form of $g(\rho)$ itself. It is convenient to reintroduce the parametrization $\delta \hat \lambda = M N \, \delta t$ and expand to second order in $\delta t$ the exponential:
\begin{align} \label{eq:marginalization_2}
	e^{u+v} &= 1 + (u+v) + \frac{u^2}{2} + O(\delta t^3) \\ \nonumber
	u &= -i \, N \delta t \, \rho_i \\ \nonumber
	v &= -\frac{N}{2} \delta t^2 \, \rho_i(1-\rho_i).
\end{align}
Since the instances are i.i.d., the marginal characteristic function is the $M$th power of this expression with $t=\lambda/(MN)$. Taking its logarithm and retaining terms through second order gives
\begin{equation} \label{eq:unconditional_characteristic}
  \log\chi_{\hat{\xi}}(\delta \hat{\lambda})
  = -i \, \delta \hat{\lambda} \, \mu - \frac{\delta \hat{\lambda}^2}{2}\left(\frac{m_2-\mu^2}{M}
  + \frac{\mathbb{E}_g[\rho(1-\rho)]}{MN}\right) + O(\delta \hat{\lambda}^3),
\end{equation}
where we have defined:
\begin{align} \label{eq:marginalization_moments}
  \mu &\equiv \mathbb{E}_g[\rho] = \int_0^1 g(\rho)\rho\,d\rho, \\
  m_2 &\equiv \mathbb{E}_g[\rho^2] = \int_0^1 g(\rho)\rho^2\,d\rho.
\end{align}
Plugging this expression back into Eq.~\ref{eq:conditional_3} we can read off the unconditional mean and variance of $\hat \xi$:
\begin{align} \label{eq:unconditional_moments}
  \mathbb{E}[\hat{\xi}] &= \mu, \\
  \operatorname{Var}(\hat{\xi})
  &= \frac{m_2-\mu^2}{M} + \frac{\mathbb{E}_g[\rho(1-\rho)]}{MN} \\ \nonumber
  &= \frac{\tau^2}{M} + \frac{\mathbb{E}_g[\sigma^2_{p_i}]}{M}.
\end{align}
The first term arises from variation in success probabilities across independently sampled instances; the second arises from the finite number of Bernoulli runs per instance. Thus the two components emerge constructively by marginalizing and expanding the characteristic function. Their sum matches Eq.~\eqref{eq:total-variance-final} in Section~\ref{sec:statistical-analysis-framework}, where the same result is presented through the law of total variance.
\section{\texorpdfstring{Finite-Sample Analysis of $pass@k$}{Finite-Sample Analysis of pass@k}} \label{ch:pass-k-cumulants}

We derive the finite-sample distribution and variance of the $pass@k$ estimator, complementing the main-text definition in Eq.~\eqref{eq:pass-at-k}. The derivation tracks both the total number of successes among $N$ runs and the outcome on a selected subset of $k$ runs.

Fix an instance $i$ and condition on its true success probability $\rho_i$. The run outcomes are exchangeable under the i.i.d. Bernoulli model, so without loss of generality we take the selected subset to be the first $k$ runs. For $1\leq k\leq N$, define
\begin{align} \label{eq:counting_terms}
  C_i &= \sum_{r=1}^{N} \xi_{ir}, \\
  D_i &= \prod_{r=1}^{k}(1-\xi_{ir}), \qquad B_i = 1-D_i.
\end{align}
Here, $C_i$ is the total number of successes, $D_i$ indicates that all selected runs fail, and $B_i$ indicates that at least one selected run succeeds. The $pass@k$ estimate conditional on the observed count $C_i=c_i$ is the probability of $B_i=1$ when a size-$k$ subset is selected uniformly at random.

The joint probability mass function of $C_i$ and $B_i$ can be written using the Kronecker delta representation $\delta_{\mathrm K}(n)=\int_{-\pi}^{\pi}\frac{d\lambda}{2\pi}e^{i\lambda n}$ for integer $n$. Define their joint characteristic function by
\begin{align} \label{eq:k_conditional_1}
  \mathbb{P}_{N,k}(c_i,b_i\mid\rho_i)
  &=\left\langle\delta_{\mathrm K}(c_i-C_i)\delta_{\mathrm K}(b_i-B_i)\right\rangle_{\xi_{ir}\mid\rho_i} \\
  &=\int_{-\pi}^{\pi}\frac{d\lambda}{2\pi}\int_{-\pi}^{\pi}\frac{dt}{2\pi}\,e^{i\lambda c_i+itb_i}\chi^{N,k}_i(\lambda,t\mid\rho_i), \\
  \chi^{N,k}_i(\lambda,t\mid\rho_i)
  &:=\left\langle e^{-i\lambda C_i-itB_i}\right\rangle_{\xi_{ir}\mid\rho_i}.
\end{align}
Both variables are discrete, which is why the Fourier integrals run over $[-\pi,\pi]$. Setting $t=0$ marginalizes over $B_i$ and recovers the characteristic function of $C_i$, namely $a_i(\lambda)^N$, where
\begin{equation}
  a_i(\lambda)=(1-\rho_i)+\rho_i e^{-i\lambda}.
\end{equation}
This is the count version of the per-instance characteristic function in Eq.~\eqref{eq:conditional_1}; the Fourier argument there is scaled by $1/N$ because it describes the sample average $C_i/N$. Since $B_i=1-D_i$ and $D_i$ is binary, we can easily see that the following identity holds:
\begin{equation}
  e^{-itB_i}=e^{-it(1-D_i)}=e^{-it}+(1-e^{-it})D_i.
\end{equation}
The first term contributes the characteristic function of $C_i$. For the second term, independence of the runs gives
\begin{align}
  \left\langle e^{-i\lambda C_i}D_i\right\rangle_{\xi_{ir}\mid\rho_i}
  &=\prod_{r=1}^{k}(1-\rho_i)\prod_{r=k+1}^{N}\left[(1-\rho_i)+\rho_i e^{-i\lambda}\right] \\
  &=(1-\rho_i)^k a_i(\lambda)^{N-k}.
\end{align}
Consequently, the joint characteristic function is
\begin{equation}
  \chi^{N,k}_i(\lambda,t\mid\rho_i)
  =e^{-it}a_i(\lambda)^N+(1-e^{-it})(1-\rho_i)^k a_i(\lambda)^{N-k}.
\end{equation}
Setting $\lambda=0$ we obtain the characteristic function of $B_i$ alone.
\subsection{\texorpdfstring{Conditional characteristic function of $B_i$}{Conditional characteristic function of B}} \label{ch:b-characteristic}
We now condition on the observed total count $C_i=c_i$. Define
\begin{align} \label{eq:conditional_int_1}
  I_m(c_i) &= \int_{-\pi}^{\pi}\frac{d\lambda}{2\pi}\,e^{i\lambda c_i}a_i(\lambda)^m, \\
  J_{N-k}(c_i) &= (1-\rho_i)^k I_{N-k}(c_i).
\end{align}
The first quantity is the probability mass of $C_i=c_i$; the second is the joint probability mass of $C_i=c_i$ and $D_i=1$. Expanding with the binomial theorem gives
\begin{equation} \label{eq:conditional_int_2}
  I_N(c_i)=\sum_{j=0}^{N}\binom{N}{j}(1-\rho_i)^{N-j}\rho_i^j
  \int_{-\pi}^{\pi}\frac{d\lambda}{2\pi}e^{i\lambda(c_i-j)}
  =\binom{N}{c_i}(1-\rho_i)^{N-c_i}\rho_i^{c_i}.
\end{equation}
Similarly,
\begin{equation} \label{eq:conditional_int_3}
  J_{N-k}(c_i)=
  \begin{cases}
    \binom{N-k}{c_i}(1-\rho_i)^{N-c_i}\rho_i^{c_i}, & c_i\leq N-k,\\
    0, & c_i>N-k.
  \end{cases}
\end{equation}
The conditional characteristic function of $B_i$ given $C_i=c_i$ is obtained by normalizing the joint Fourier coefficient by $I_N(c_i)$:
\begin{equation} \label{eq:conditional_characteristic_B}
  H_{i,c_i}(t)
  :=\mathbb{E}[e^{-itB_i}\mid C_i=c_i,\rho_i]
  =\frac{1}{I_N(c_i)}\int_{-\pi}^{\pi}\frac{d\lambda}{2\pi}\,e^{i\lambda c_i}\chi^{N,k}_i(\lambda,t\mid\rho_i).
\end{equation}
Substituting the expressions for $I_N$ and $J_{N-k}$ yields
\begin{equation} \label{eq:conditional_characteristic_B_final}
  H_{i,c_i}(t)=e^{-it}+(1-e^{-it})\frac{J_{N-k}(c_i)}{I_N(c_i)}.
\end{equation}
The ratio $J_{N-k}(c_i)/I_N(c_i)$ is the conditional probability that all $k$ selected runs fail. Therefore, the conditional mean of $B_i$—and hence the $pass@k$ estimate—is
\begin{equation}
  h_k(c_i)=i\left.\frac{\partial}{\partial t}\log H_{i,c_i}(t)\right|_{t=0}
  =1-\frac{J_{N-k}(c_i)}{I_N(c_i)}
  =1-\frac{\binom{N-c_i}{k}}{\binom{N}{k}},
\end{equation}
where $\binom{N-c_i}{k}=0$ when $N-c_i<k$. Thus the estimator for instance $i$ is $Q_{i,k}=h_k(C_i)$.

\subsection{Sampling variance} \label{sec:sampling-variance}
At fixed $\rho_i$, the count $C_i$ has probability mass function $I_N(c_i)$. The characteristic function of the estimator $Q_{i,k}$ is therefore
\begin{align}
  \Psi_{i,k}(u\mid\rho_i)
  &=\mathbb{E}[e^{-iuQ_{i,k}}\mid\rho_i]
  =\sum_{c_i=0}^{N}I_N(c_i)e^{-iu h_k(c_i)}, \\
  \Phi_{i,k}(u\mid\rho_i)&=\log\Psi_{i,k}(u\mid\rho_i).
\end{align}
With the characteristic-function convention used here, the first derivative of $\Phi_{i,k}$ at zero is $-i\mathbb{E}[Q_{i,k}\mid\rho_i]$, and the second is $-\operatorname{Var}(Q_{i,k}\mid\rho_i)$:
\begin{align}
  \Phi_{i,k}'(0\mid\rho_i)&=-i\sum_{c_i=0}^{N}I_N(c_i)h_k(c_i), \\
  \Phi_{i,k}''(0\mid\rho_i)&=-\sum_{c_i=0}^{N}I_N(c_i)h_k(c_i)^2
  +\left[\sum_{c_i=0}^{N}I_N(c_i)h_k(c_i)\right]^2.
\end{align}
The conditional mean simplifies to
\begin{align}
  \mu_{i,k}:=\mathbb{E}[Q_{i,k}\mid\rho_i]
  &=1-\sum_{c_i=0}^{N-k}\binom{N-k}{c_i}(1-\rho_i)^{N-c_i}\rho_i^{c_i} \\
  &=1-(1-\rho_i)^k.
\end{align}
The conditional sampling variance is
\begin{equation} \label{eq:var-hk-rho-appendix}
  v_{i,k}:=\operatorname{Var}(Q_{i,k}\mid\rho_i)
  =\sum_{c_i=0}^{N}I_N(c_i)h_k(c_i)^2-\left[1-(1-\rho_i)^k\right]^2.
\end{equation}
This is the variance formula reported in the main text. Averaging these conditional moments over the distribution of $\rho_i$ gives the corresponding unconditional moments.

\subsection{Simple identities} \label{ch:simple-identities}

For $k=1$, the finite-sample formula reduces to the empirical pass rate. For $C_i<N$, expanding the binomial coefficients gives
\begin{align}
  h_1(C_i)
  &=1-\frac{\binom{N-C_i}{1}}{\binom{N}{1}} = 1-\frac{(N-C_i)!}{(N-C_i-1)!}\,\frac{(N-1)!}{N!} \\ \nonumber
  &=1-\frac{N-C_i}{N}
  =\frac{C_i}{N}
  =p_i.
\end{align}
For $C_i=N$, the numerator binomial coefficient is zero, so the same identity gives $h_1(N)=1=p_i$.
For the large-$N$ approximation, write $p_i=C_i/N$ and use the product representation of the binomial-coefficient ratio:
\begin{equation}
  h_k(C_i)=1-\prod_{j=0}^{k-1}\frac{N-C_i-j}{N-j}
  =1-\prod_{j=0}^{k-1}\frac{1-p_i-j/N}{1-j/N}
  \simeq 1-(1-p_i)^k,
\end{equation}
where the approximation holds as $N$ grows with $k$ fixed (equivalently, when $N\gg k$).

\section{\texorpdfstring{Moments of the Plug-in $\passhatk$ Estimator}{Moments of the Plug-in pass\textasciicircum k Estimator}} \label{ch:pass-hat-k-plugin-moments}
For a fixed task, let $C_i\mid\rho_i\sim\operatorname{Binomial}(N,\rho_i)$ and $\widetilde Q_k(C_i)=(C_i/N)^k$. To express its exact moments, let $(N)_j=N(N-1)\cdots(N-j+1)$ be the falling factorial and let $\left\{ {m \atop j} \right\}$ be a Stirling number of the second kind. Using the identities
\begin{equation}
  C_i^m=\sum_{j=1}^{m}\left\{ {m \atop j} \right\}(C_i)_j,
  \qquad
  \mathbb{E}[(C_i)_j\mid\rho_i]=(N)_j\rho_i^j,
\end{equation}
with $(N)_j=0$ for $j>N$, the exact conditional mean is
\begin{equation} \label{eq:pass-hat-k-plugin-mean}
  \mathbb{E}[\widetilde Q_k(C_i)\mid\rho_i] = \frac{1}{N^k} \sum_{j=1}^{k}
  \left\{ {k \atop j} \right\}(N)_j\rho_i^j.
\end{equation}
For $k>1$ and $0<\rho_i<1$, this exceeds $\rho_i^k$, whereas the unbiased estimator in Eq.~\eqref{eq:pass-hat-k-unbiased} has conditional mean exactly $\rho_i^k$. Applying the same identity to the second moment gives the exact conditional variance
\begin{align} \label{eq:pass-hat-k-plugin-var}
  \operatorname{Var}(\widetilde Q_k(C_i)\mid\rho_i) &=\frac{1}{N^{2k}} \sum_{j=1}^{2k}
  \left\{ {2k \atop j} \right\}(N)_j\rho_i^j -
  \left[
    \frac{1}{N^k}
    \sum_{j=1}^{k}
    \left\{ {k \atop j} \right\}(N)_j\rho_i^j
  \right]^2.
\end{align}
Its square root is the conditional standard error of the plug-in point estimator. Replacing $\rho_i$ by $p_i=C_i/N$ gives a plug-in estimate of this standard error, as discussed in the main text.
\section{Choosing the sample size} \label{ch:scaling}
In this section, we assess how the scenario-level pass-rate estimate and its uncertainty depend on the number of runs available per instance.
\nbfigure[0.8\textwidth][!tb]{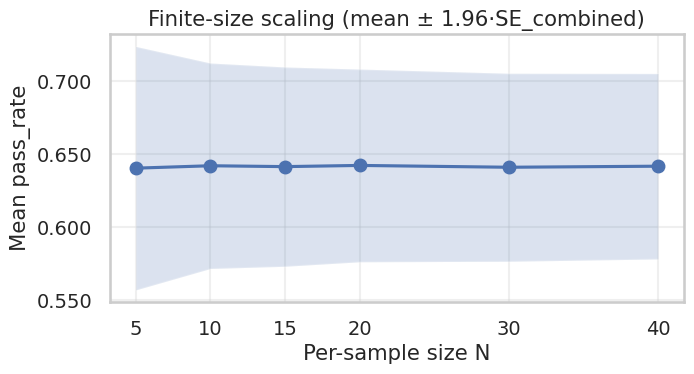}{
  Scenario-level pass-rate estimate as a function of the number $n$ of runs per instance. The curve shows the mean across subsampling replicas, and the shaded band shows approximate 95\% intervals based on the hierarchical standard error from Eq.~\eqref{eq:total-variance-final}.
}{finite_size_scaling}
For each instance, we have up to $N_{\max}$ runs (here, $N_{\max}=40$). Fix a scenario containing $M$ instances, and let $n\leq N_{\max}$ be the target number of runs per instance; each included instance must have at least $n$ observed runs. For subsampling replica $b$, draw a set $\mathcal{R}_{i,b}(n)$ of $n$ distinct run indices for each instance $i$, and compute
\begin{equation}
  p_{i,b}(n)=\frac{1}{n}\sum_{r\in\mathcal{R}_{i,b}(n)}\xi_{ir},
  \qquad
  \bar{p}_b(n)=\frac{1}{M}\sum_{i=1}^{M}p_{i,b}(n).
\end{equation}
The draws are without replacement within each replica, so no observed run is counted twice in that replica; different replicas may overlap. Since the original runs are i.i.d. conditional on each instance's success probability, each such subset has the same marginal sampling distribution as $n$ runs from that instance. We repeat the subsampling with $R$ random seeds and report the replica-averaged scenario pass rate
\begin{equation}
  \widehat{\bar{p}}(n)=\frac{1}{R}\sum_{b=1}^{R}\bar{p}_b(n).
\end{equation}
Figure~\ref{fig:finite_size_scaling} shows this mean and approximate 95\% intervals, $\widehat{\bar{p}}(n)\pm1.96\,\widehat{\operatorname{SE}}_{\mathrm{hier}}(n)$, using the hierarchical variance propagation in Eq.~\eqref{eq:total-variance-final} with $N=n$. At the full sample size, $n=N_{\max}$, no subsampling is performed. As $n$ increases, the within-instance binomial-noise contribution decreases, while between-instance variability remains; accordingly, the aggregate uncertainty does not in general vanish as $n$ grows. The binomial-noise contribution scales as $1/(Mn)$, so increasing the number of runs has diminishing returns, particularly when $M$ is large. The normal-approximation intervals should be interpreted cautiously for small evaluation sets, where they can understate uncertainty~\cite{bowyer2025}. This analysis illustrates the trade-off between computational cost and the statistical precision of scenario-level performance estimates.

\end{document}